\documentclass[review]{elsarticle}

\usepackage{lineno}
\usepackage[]{hyperref}
\usepackage{bm}
\usepackage{times}
\usepackage{epsfig}
\usepackage{graphicx}
\usepackage{amsmath}
\usepackage{amssymb}
\usepackage{tabularx} 
\newcolumntype{Y}{>{\centering\arraybackslash}X}

\DeclareMathOperator{\Tr}{Tr}
\usepackage{array,booktabs,arydshln,xcolor} 
\usepackage{epstopdf}
\usepackage{algorithm}
\usepackage{amsthm}
\usepackage{multirow}
\usepackage[algo2e]{algorithm2e}
\usepackage[11pt]{moresize}
\usepackage{tabu}
\usepackage{caption}
\usepackage{subcaption}
\usepackage{threeparttable}

\journal{Pattern Recognition}

\begin{document}

\begin{frontmatter}

\title{Revisiting Graph Construction for Fast Image Segmentation}

\author{Zizhao Zhang$^1$, ~Fuyong Xing$^2$, Hanzi Wang$^4$, \\ Yan Yan$^4$, ~Ying Huang$^4$,  ~Xiaoshuang Shi$^{3}$, ~Lin Yang$^{1，2，3,\star}$ }
\address{$^1$Dept. of Computer and Information Science and Engineering, University of Florida, FL 32611, USA \\
	$^2$Dept. of Biostatistics and Informatics, Colorado School of Public Health, University of Colorado Denver, Denver, CO 80045 USA \\
	$^3$J. Crayton Pruitt Family Dept. of Biomedical Engineering, University of Florida, FL 32611, USA \\
	$^4$  Fujian Key Laboratory of Sensing and Computing for Smart City, School of
	Information Science and Engineering, Xiamen University, Fujian 361005, China }


\cortext[mycorrespondingauthor]{Corresponding author}
\ead{zizhaozhang@ufl.edu, fuyong.xing@ucdenver.edu, hanzi.wang@xmu.edu.cn, yanyan@xmu.edu.cn, yinghuang@stu.xmu.edu.cn， xsshi2013@gmail.com, lin.yang@bme.ufl.edu }


\begin{abstract}
In this paper, we propose a simple but effective method for fast image segmentation. We re-examine the locality-preserving character of spectral clustering by constructing a graph over image regions with both global and local connections. Our novel approach to build graph connections relies on two key observations: 1) local region pairs that co-occur frequently will have a high probability to reside on a common object; 2) spatially distant regions in a common object often exhibit similar visual saliency, which implies their neighborship in a manifold. We present a novel energy function to efficiently conduct graph partitioning. Based on multiple high quality partitions, we show that the generated eigenvector histogram based representation can automatically drive effective unary potentials for a hierarchical random field model to produce multi-class segmentation. Sufficient experiments, on the BSDS500 benchmark, large-scale PASCAL VOC and COCO datasets, demonstrate the competitive segmentation accuracy and significantly improved efficiency of our proposed method compared with other state of the arts. 
\end{abstract}


\end{frontmatter}


\section{Introduction}
Image segmentation is a challenging and critical computer vision task. Graph-based algorithms have been shown as an effective approach for image segmentation \cite{felzenszwalb2004efficient,boykov2001fast,peng2013survey}. Among various graph based approaches, spectral clustering becomes a major trend \cite{kim2013learning,shi2015framework}. 

Recent methods attempt to solve several primary issues of spectral clustering (referring to normalized cuts (NCut) \cite{shi2000normalized}) based image segmentation to segment image into meaningful partitions. First, NCut based methods tend to segment image into spatially connected components \cite{shi2000normalized,yu2015piecewise}. Multiscaling processing \cite{yu2005segmentation,maire2013progressive} is a common way to address this problem by building the affinity for distant pixel affinities \cite{cour2005spectral,maire2013progressive}. 
However, the usage of these methods for real large-scale datasets is not clear. Most current cutting-edge methods do not follow this direction. Instead, recent methods, like gPb \cite{arbelaez2011contour} and MCG \cite{arbelaez2014multiscale,yu2015piecewise} based methods \cite{chenscale} use the boundary-preserving property of NCut to trace boundary orientation information rather than direct segmentation. Building effective affinity matrices \cite{yu2015piecewise,kim2013learning,arbelaez2014multiscale} usually uses sophisticated low-level features \cite{arbelaez2011contour}. These features can effectively measure the local changes but are not effective in capturing high-level knowledge for segmentation. They are not good options for fast segmentation either due to high computational cost \cite{arbelaez2011contour}. Different from previous approaches, our method re-examines spectral clustering from a manifold learning perspective to construct a graph to model the high-level image knowledge (i.e., pixel pair co-occurrence and saliency relationship) for unsupervised image segmentation. More importantly, our method provides the possibility of enabling graph partitioning to directly segment challenging natural images rather than just boundary tracing.

\begin{figure*}[t]
	\begin{center}
		
		\includegraphics[width=0.9999\linewidth]{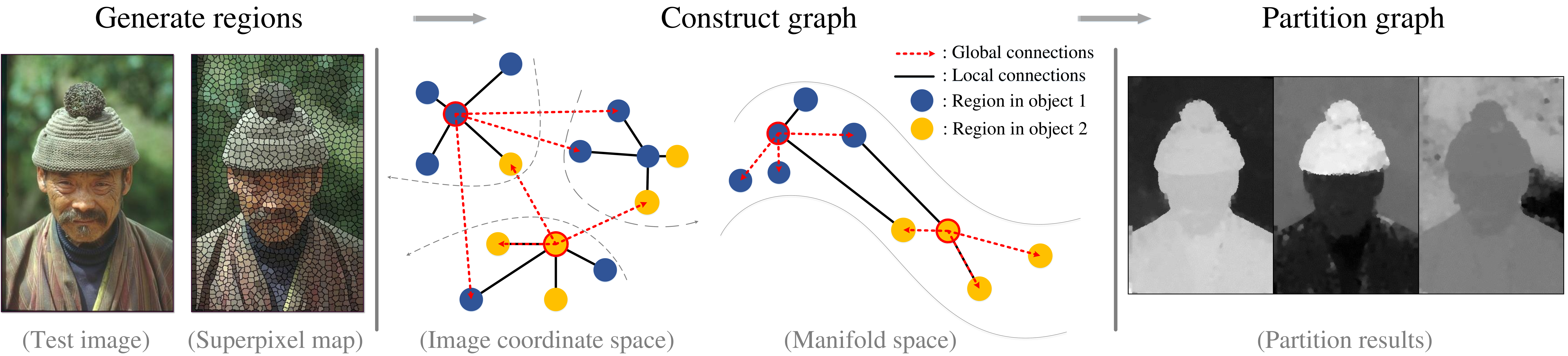}
		
	\end{center}

	\caption{Illustration of the graph construction and partition procedures. The test image is first divided into over-segmented regions (left). Each region is treated as a graph node (middle). Local connections (black solid lines) link each region to its spatially adjacent regions in the image plane. When viewing those regions in a certain manifold space, spatially distant regions in the same object will be adjacent; directed global connections (red dotted arrows) link each region to its neighbors. We cut the graph to obtain multiple partitions (right). }  \label{fig:outline}
\end{figure*}

To better illustrate the motivation, we first explain the latent relation of NCut to manifold learning. Both NCut and Laplacian eigenmaps \cite{belkin2003laplacian} take advantage of the locality-preserving character \cite{ng2002spectral}  of graph Laplacian to conduct clustering and dimensionality reduction. In fact, locality-based dimensionality reduction methods are implicitly tied to clustering \cite{belkin2003laplacian,shi2014face}. Preserving locality is the key factor that drives effective clustering. Let's assume pixels of an image lie on a certain manifold where pixels belonging to a common object are adjacent (within a small range), but far away in the spatial image plane. These pixels supposed to have strong connections to be grouped together, but these connections are not encoded in the sparse affinity matrix of NCut due to their Euclidean distances. Although multi-scale affinity matrices \cite{yu2005segmentation,cour2005spectral} can alleviate this issue, increasing the range of an affinity matrix and connecting all pixel pairs in the range also introduce unavoidable noises. The method to construct affinities between spatially distant and adjacent pixels should be considered respectively in order to better capture their respective characteristics in image statistics.

In this paper, we propose a novel approach to construct an image region graph to address the aforementioned problems. The overall idea is illustrated in Figure \ref{fig:outline}.
The graph nodes are connected among both spatially adjacent and distant regions through different and independent cues. We build local connections between spatially adjacent regions with an affinity matrix. The estimation of the similarity between two regions is based on an observation that adjacent region pairs co-occurring frequently often reside on a common object. Oppositely, global connections are built among adjacent regions in the manifold which might be spatially distant, with an objective to preserve their relationships and encourage them to be clustered together. We introduce a simple cue to discover the similar saliency of those regions as the global connection measurement. We present a new energy function to partition the constructed graph, which formulates the minimization problem as a single and efficiently solvable eigenvector system.  Based on the generated high quality graph partitions, we present a simple eigenvector histogram based representation to represent image regions and automatically drive effective unary potentials for the hierarchical random field of the Pylon model \cite{lempitsky2011pylon}, yielding high-quality multi-class segmentation.

In brief, the contributions of this paper are:
\begin{itemize}
	\itemsep-.2em
	\item We propose a sophisticatedly connected graph to build the connection of image regions yet with very efficient graph partitioning capability.
	\item We exploit various simple and efficient cues to capture the high-level image information in order to segment objects with complex inner-variances and background.
	\item We present a multi-class segmentation strategy by utilizing graph partitions to generate clear and smooth segmentation. 
	\item Extensive experiments and comprehensive analysis are conducted, on BSDS500 \cite{arbelaez2011contour}, large-scale PASCAL VOC \cite{everingham2010pascal} and COCO \cite{lin2014microsoft} datasets, to validate the effectiveness of our proposed method, its generalization ability to different datasets with diverse scenes, and the high efficiency compared with other state of the arts.
\end{itemize}

The rest of the paper are structured as follow: Section 2 discusses the related work. Section 3 introduces the graph construction and partition of our method. Section 4 introduces the proposed multi-class segmentation by utilizing graph partitions. Finally, Section 5 conducts experiments and detailed analysis. Section 6 concludes the paper.

\section{Related Works}
Image segmentation has been studied in the computer vision community for decades. Shi \textit{et al.} \cite{shi2000normalized} propose normalized cuts (NCut), which advanced spectral clustering based image region segmentation. \cite{yu2003multiclass} enables its multi-class segmentation.  Among the region based segmentation, diffusion based approaches \cite{wang2012affinity,zhang2010diffusion}, GraphCut \cite{rother2004grabcut}, GrabCut \cite{rother2004grabcut}, etc \cite{grady2006random,li2016multiscale,yin2017unsupervised,de2014graph}, have been explored to partition images. Building successful affinity matrices is critical \cite{zhu2014constructing}.
Many subsequent approaches have computed more effective affinity matrices using elaborately designed low-level features and metrics \cite{kim2013learning,cour2005spectral,arbelaez2011contour,li2012segmentation}. To solve the limitation of NCut to capture affinities of distant pixels, several methods \cite{yu2005segmentation,maire2013progressive,li2012segmentation,wang2015global} have been proposed base on multi-scaling affinity strategies. However, dense affinity suffers from optimization bottleneck, although approximation algorithms are explored \cite{cour2005spectral,maire2013progressive,arbelaez2014multiscale}. Our method is able to capture both local and global affinities as well keeps the sparsity of the affinity matrix.

Contour driven image region segmentation is widely studied. Arbelaez \textit{et al.} \cite{arbelaez2011contour} propose the globalized probability of boundary (gPb), which utilizes the boundary-preserving characteristic of NCut with sophisticatedly designed features to detect object boundaries and incorporate it into the oriented watershed transform and ultrametric contour map (OWT-UCM) to conduct image segmentation. This approach becomes the main support of many subsequent segmentation approaches \cite{isola2014crisp,kim2013learning,xiaofeng2012discriminatively,donoser2014discrete,arbelaez2014multiscale,chenscale}. Kim \textit{et al.} \cite{kim2014image} formulate a hypergraph-based model and perform correlation clustering for image segmentation. Recently, Yu \textit{et al.} \cite{yu2015piecewise} minimize an \(\ell_{1}\)-normed energy function of NCut to obtain piecewise smooth embeddings for gPb-owt-ucm \cite{arbelaez2011contour}, which obtains state-of-the-art image segmentation performance. However, all these methods suffer from expensive computations for feature extraction or optimization. Speed issues are considered in several following work.
Multiscale combinatorial grouping (MCG) \cite{arbelaez2014multiscale} segment images with multi-scale UCMs and it uses more advanced edge detection methods \cite{dollar2015pami} to largely reduce the computation bottleneck of NCut used by gPb-owt-ucm. Chen \cite{chenscale} provides a solution to the scale-alignment in MCG.  However, all these methods suffer from expensive computations for feature extraction or optimization. Sometimes several minutes are required to process a single $321 \times 481$ image, which significantly limits their practical usages.
On the other hand, Pont-Tuset \textit{et al.} \cite{pont2017multiscale} propose a downsampled approximating algorithm to accelerate the graph partitioning and use richer information in multiscale UCMs.
Tayor \textit{et al.} \cite{taylor2013towards} and many others \cite{kim2013learning,kim2014image} reduce the size of the affinity matrix using superpixel techniques. In this paper, we develop a method that is much faster than the aforementioned methods with competitive accuracy. 

Edge detection plays an extreme importantly role in region based image segmentation \cite{shen2017multi,Man+17, shen2015deepcontour,zhang2016semicontour,shen2016object}. For example, Convolutional Oriented Boundaries (COB) proposes an accurate boundary detection method using convolutional neural networks (CNNs) and combines with  \cite{pont2017multiscale} to perform image and object segmentation.  Another popular image segmentation direction is semantic segmentation. Current methods use CNNs \cite{long2015fully,chen2016deeplab,noh2015learning} to predict the semantic label of each pixel. These methods rely on large-scale training data. In contrast, our method aims at partition images into regions that can accurately segment objects from an image by observing its internal statistics in an unsupervised manner.

Designing feature to build the affinity between pixels/regions is important. Several studies have explored different cues, such as sophisticated combination of mixed image features \cite{arbelaez2011contour}, texture information \cite{zhou2013texture}, or saliency \cite{LI2016317}. Different from these low-level image features, we argue that high-level cues are equally important and sometimes even more effective. For example, 
co-occurrence statistics have been used to capture the semantic object context knowledge based on training data to help the inference in, for example, condition random field (CRF) \cite{ladicky2010graph}. Different from this direction of research, our approach models region-wise co-occurrence probability based on pointwise mutual information \cite{fano1961transmission} to build local connections of our proposed graph learned from the image itself.

Laplacian eigenmaps \cite{belkin2001laplacian} computes a low-dimensional embedding to preserve the pairwise affinity of data points in the manifold. Local linear embedding (LLE) \cite{roweis2000nonlinear}, alternatively, preserves the linear structure among the local neighboring points. The locality-preserving character of these two methods implicitly encourages the clustering of data. However, Isomap \cite{tenenbaum2000global}, which preserves global data geodesic distances, does not possess the nature of clustering. Our method shows a distinct point of view on the side of manifold learning to enhance spectral clustering for image segmentation.

\section{Global-local Connected Graph Partitioning}
\label{sec:eigenv}
In this section, we present the approach to build the local and global connections of the graph. Then we introduce the proposed energy function to partition the constructed graph. 

\subsection{Local connection with co-occurrence cues}
Our proposed method begins with an over-segmentation with a set of regions, defined as $\mathcal{S}=\{S_1,...,S_N\}$. The over-segmentation is favorable considering its local spatial consistency and computational efficiency. 
Denote the graph by $\mathcal{G}_{local} = \{\mathcal{S}, W\}$, where $W \in \mathbb{R}^{N\times N}$ is the affinity matrix with each entry $W_{ij}$ representing the affinity between regions $S_i$ and $S_j$. $W$ is sparse such that only spatially adjacent region pairs within a small range have nonzero values. Given a test image with large appearance variations inside the object (see Figure \ref{fig:eigventors}), a desirable affinity matrix should be able to discover the strong affinity between two visually different neighboring regions belonging to the common object. However, it is difficult for low-level features to achieve this goal because of their limitations in learning high-level knowledge.

One type of high-level knowledge comes from the fact that a neighboring region pair residing on an object is more likely to co-occur (i.e., have a high joint probability) due to the color patterns inside the object \cite{isola2014crisp}, such as the strip patten on the clothes of the images in Figure \ref{fig:eigventors}. If we treat regions $\{S_i|i=1,...,N\}$ as random variables, we can define the co-occurrence of two regions as

\begin{equation}
\label{eq:cooccur}
CO(S_i, S_j) = \log \frac{1}{A} P(S_i, S_j), 
\end{equation}
where $P(S_i, S_j)$ is the joint probability over $S_j$ and $S_j$. Let $A = P(S_i)P(S_j)$ represent a normalization term, which is crucial to penalize the biased-high $P(S_i, S_j)$ of background region pairs against foreground object region pairs, because the background area usually has larger proportion than foreground objects. This normalization term will eliminate this unbalance accordingly.
In addition, $CO$ also contains information about object boundaries, because a region pair across the object boundary is a small-probability event \cite{isola2014crisp}.

We estimate $P(S_i, S_j)$ and marginal distribution $P(S_i)$ by using a nonparametric kernel density estimator \cite{parzen1962estimation} following \cite{isola2014crisp}. But differently, we densely sample region pairs of each region and its adjacent regions within a certain (denoted as $e_1$) distance apart without repetition (which means $P(S_1, S_2) =  P(S_2, S_1)$).
Basically, we place estimator kernels on all regions $\{S_i\}$, and compute the image feature (gray values) co-occurrence probability over all region pairs. So for each feature value pair, we have a co-occurrence frequency. Then we can simply normalize them and obtain $P(S_i)$  and the final co-occurrence cue $CO(S_i, S_j)$.

Our approach shares some similarities with \cite{isola2014crisp} (denoted as PMI) for using pointwise mutual information, but is different from PMI in several perspectives. PMI interests in low pixel-wise joint probability to discover the rare boundaries, but we are interested in high region-wise probabilities and simultaneously maintain the boundary detection ability of PMI. PMI relies on raw image pixels, the probability in Eq. (\ref{eq:cooccur}) is estimated over limited number of randomly sampled pixels. We rely on coherent regions to estimate this probability over most of adjacent region pairs, which yields probability distribution estimation closer to the actual distribution for the regions, and the estimation process is much less computational expensive.

\vspace{0.1cm}
\noindent
\textbf{Energy function:} The first term $E_{local}$ in our proposed energy function will encourage frequently co-occurring region pairs to be clustered into a group, and vice versa. Minimizing $E_{local}$ is defined as the following:

\begin{equation}
\label{eq:ncut}
\min_{\bm y}\;\,  \sum_{i=1}^{N} \sum_{j=1}^{N}||y_i - y_j||^2 W_{ij}, \quad s.t. \; {\bm y}{^T}D\bm{y}=1, 
\end{equation}
where $D$ is a diagonal matrix and its $i$-th diagonal element is $d_{ii} = \sum_j W_{ij}$. The constraint is the key to normalize the cut of the graph. Minimizing $E_{local}$ enforces $y_i$ and $y_j$ to take a similar value when $W_{ij}$ is large.
$\bm{y} = [y_1,...,y_N]^T$ is a real-valued vector, which is interpreted as a binary graph partition in NCut or an one dimensional embedding in Laplacian eigenmaps. $W_{ij}$ is defined as
\begin{equation}
W_{ij} = \exp{\left(\sum_{o} CO(S^{f_o}_i, S^{f_o}_j)\right)},
\label{eq:affmat}
\end{equation}
where the superscript $f_o$ specifies a feature representation of the corresponding region. For each region, we calculate the pixel mean of Lab color space and the diagonal values of the RGB color covariance matrix in a $3\times 3$ window. $W_{ij}$ is computed between $S_i$ and $S_j$ within a certain distance apart, denoted as $e_2$ ($e_2 > e_1$).

The affinity matrix $W$ is designed to measure the similarity between spatially adjacent region pairs based on their latent co-occurrence statistics. In order to preserve the ignored relationships among spatially distant regions in the common object, we propose an additional energy term by building the global connections of the graph in the following section.

\subsection{Global connection with saliency cues}
\label{sec:global}
The graph associated with global connections is denoted by $\mathcal{G}_{global}=\{\mathcal{S}, K\}$. Our approach strengthens the locality-preserving character by discovering the underlying linear structures among spatially distant regions (i.e., each region can be linearly represented by several neighboring regions so that the global connections are directed) belonging to a common object, while these regions are adjacent on a certain manifold. This goal is achieved by minimizing the second energy term $E_{global}$:

\begin{equation}
\label{eq:elocal}
\min_{\bm y}\;\, \sum_{i=1}^{N} ||y_i - \sum_{j \neq i} K_{ij} y_j ||^2, \quad s.t. \; {\bm y}^T\bm{y} = 1,  
\end{equation}
where $K \in \mathbb{R}^{N\times N}$ is the coefficient matrix with $R$ non-zero entries in each row to specify the linear combination coefficients of the representing neighbors. The constraint avoids degenerated solutions. ${\bm y}$ is interpreted as an embedding in the original locally linear embedding (LLE) \cite{roweis2000nonlinear} method.  Note that both $\bm y$ and $K$ are unknown; minimizing this energy function consists of three steps: 1) finding $R$ neighbors for each region, 2) computing coefficient matrix $K$, and 3) computing $\bm y$.

\vspace{.2cm}
\noindent
\textbf{Geodesic distance based neighbors:} For each region, we consider its candidate neighbors from all regions within a large range of the defined geodesic distance, such that the distance ($s_{ij}$) between regions $S_i$ and $S_j$ is defined as follows:
\begin{equation}
\begin{split}
s_{ij} & = \big(\min(|w_{S_i} - w_{S_j}|, I_w-|w_{S_i}-w_{S_j}|)^2 \\
&+ \min(|h_{S_i} - h_{S_j}|, I_h-|h_{S_i}-h_{S_j}|)^2\big)^{\frac{1}{2}},
\end{split}
\label{eq:geodist}
\end{equation}
where $w_{S_i}$ and $h_{S_i}$ denote the spatial $x$- and $y$-coordinates of region $S_i$, respectively. $I_w$ and $I_h$ denote the width and height of the image, respectively. Intuitively, this metric treats the image as if it was wrapped along its four corners into a sphere and describes the geodesic distance along this resulting surface. The measurement can trace the connections of the regions belonging to foreground objects or background with an arbitrary shape and range.

For a region $S_i$, we select $R$ nearest regions with each region represented as a feature vector calculated by a saliency cue mapping $\sigma$. Then we find its coefficients $K_i$ by
\begin{equation}
\min_{K_i}\;\, ||\sigma(S_i) - \sum_{j \neq i} K_{ij} \sigma(S_j) ||^2  +  \alpha \Tr(K_i^TK_i), \;\; s.t. \; \sum_{j} K_{ij}=1. 
\label{eq:Kmat}
\end{equation}
The regularization term is necessary to prevent ill-conditioned solutions when neighboring regions have similar feature values (i.e., making the Gram matrix singular). The regularization parameter is chosen as $\alpha = 1e{-}10$. The constraint ensures the translation invariance. 
\vspace{.1cm}

\begin{figure}[t]
	\begin{center}
		\includegraphics[width=0.999\linewidth]{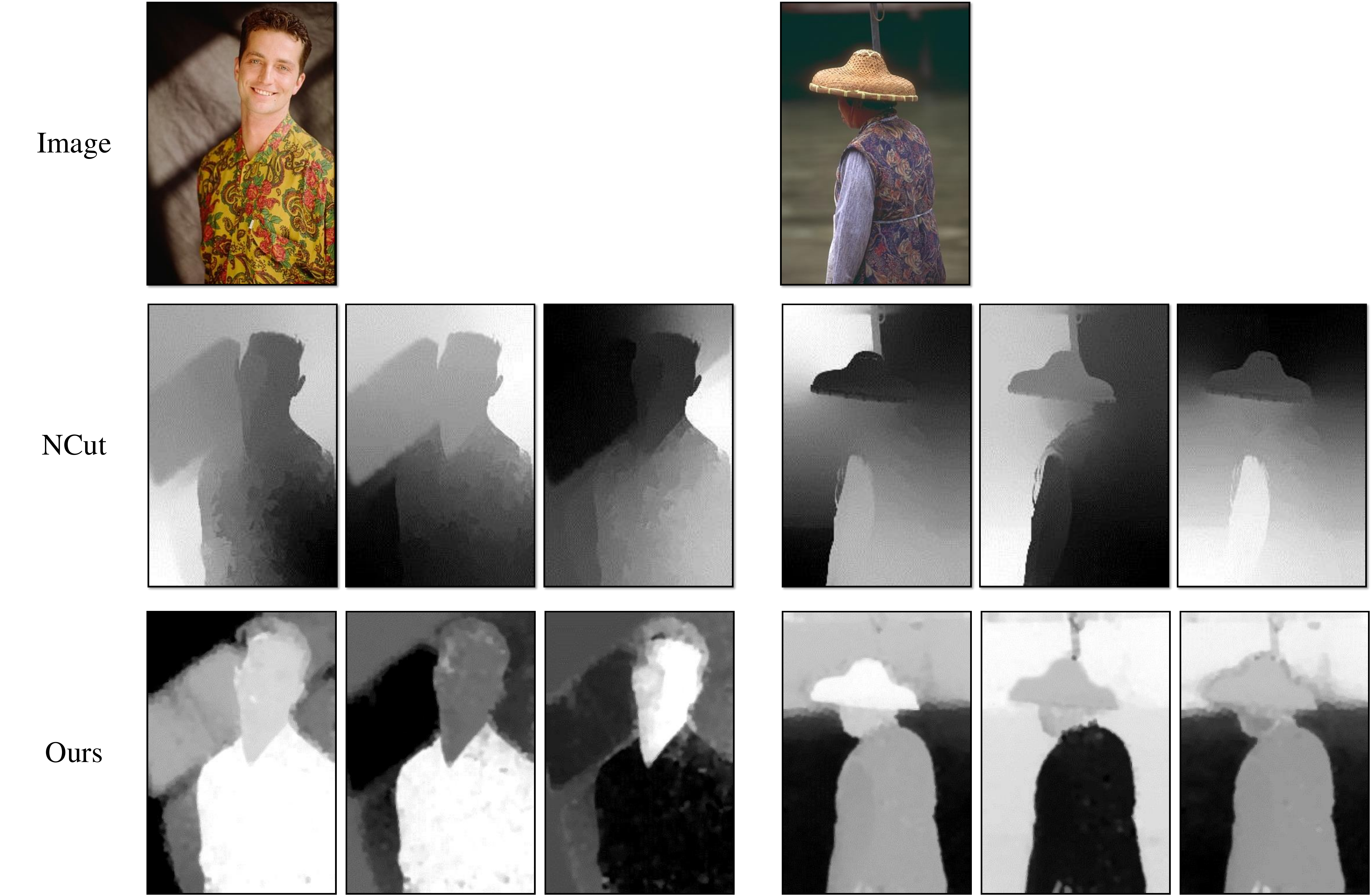}
	\end{center}

	\caption{The first row shows two sample images exhibiting large object internal  appearance variances. The second and third row shows the graph partitioning results of NCut and ours, respectively. As can be observed, NCut preserves the edge information but segments images into the connected components (middle). Our approach is able to separate the entire object from its background (bottom). }

	\label{fig:eigventors}
\end{figure}

\noindent
\textbf{Saliency cue complying linearity:} Spatially distant regions inside the same object may have large appearance variances, for example, the face, hairs, and clothes of a person exhibit totally different appearances (see Figure \ref{fig:eigventors}). Therefore, it is difficult to measure their latent similarity with traditional cues. 
However, those visually different regions usually exhibit similar saliency degree in the human visual system \cite{cheng2015global}. This characteristic remedies the ``imperfection" of pairwise co-occurrence affinity and satisfies the requirement to build global connections.
We take advantage of the empirical knowledge that salient objects in images have distinctive colors from the background under a certain linear combination of mixed color spaces \cite{kim2014salient}. To this end, we choose RGB, Lab, and hue and saturation channels of HSV (8 channels) and their nonlinear transformations with gamma correction (with three gamma values, $[0.5,1.5,2.0]$) to consider the human vision's nonlinear responses, thereby yielding a 24-dimensional feature vector for each region, $\sigma: \mathcal{S} \mapsto \mathbb{R}^{24}$.

Our method to incorporate saliency in the graph connection is elaborate. Unlike saliency detector \cite{kim2014salient}, we do not compute the coefficient explicitly based on any supervised information. Since the correlation of each region feature vector $\sigma(S)$ is consistent under arbitrary linear transformation, its saliency characteristic between regions will be implicitly expressed in Eq. (\ref{eq:elocal}). 

\subsection{Proposed energy function to partition graph}
Overall, the proposed full graph is defined as $\mathcal{G}=\{\mathcal{S}, (W, K)\}$, where $W$ specifies the local undirected connections and $K$ specifies the global directed connections. Our goal is to pursue global partitioning of the graph $\mathcal{G}$, i.e., minimizing the two energy terms simultaneously. Therefore, the energy function $E$ can be defined and derived as follows (detailed derivations are skipped):
\begin{equation}
\label{eq:energy}
\begin{split}
E &= E_{local} + \mu E_{global} \\
& = \sum_{i=1}^{N} \sum_{j=1}^{N}||y_i - y_j||^2 W_{ij} + \mu \sum_{i=1}^{N} ||y_i - \sum_{i \neq j} K_{ij} y_j ||^2 \\
& = {\bm y}^T(D-W + \mu M) \bm{y}, 
\end{split}
\end{equation}
where $(D-W) \in \mathcal{R}^{N\times N}$ is the Laplacian matrix and $M = (I - K)^T(I-K) \in \mathcal{R}^{N\times N}$. $\mu$ is a regularization parameter to balance $E_{global}$ and $E_{local}$. How to select the optimal value of $\mu$ is discussed in the experimental section.

It is straightforward to see that minimizing $E$ is to solve a generalized eigenvector system:
\begin{equation}
\label{eq:geigen}
(D-W + \mu M) \bm{y} = \lambda D\bm{y},
\end{equation}
which produces a set of eigenvectors $\hat{Y}$, where each column is an eigenvector representing a binary partition of the graph. 
In practice, the number of segments of an arbitrary test image is unknown and the expected partition is not guaranteed to be the eigenvector associated with the second smallest eigenvalue \cite{shi2000normalized}.
In Section \ref{sec:EigenHistogram}, we present a novel approach to address this issue for multi-class segmentation.
%

\vspace{.1cm}
\noindent
\textbf{Leverage $E_{local}$ and $E_{gloal}$:} The two energy terms are designed for different purposes. $E_{local}$ preserves the pairwise similarity of spatially adjacent region pairs, while $E_{global}$ preserves the linear structure of spatially distant regions in the common object. Both emphasize the locality-preserving for the purpose of clustering (or graph partitioning). Compared with the hard constraint of $E_{local}$, $E_{global}$ encourages soft (i.e., likelihood) clustering of the regions \cite{belkin2003laplacian}. In Figure \ref{fig:eigventors}, we visualize several graph partition results of the proposed approach and compare it to NCut. As can be observed, the significantly improved graph partitioning quality demonstrates the effectiveness of the global connections introduced in $E_{global}$.

\section{Multi-class Segmentation}
\label{sec:EigenHistogram}
In this section, we introduce the approach to use graph partitions for multi-class segmentation.
\subsection{EigenHistogram}
We have computed a set of eigenvectors (i.e., image partitions) $\hat{Y} = [\hat{\bm y}_1,..., \hat{\bm y}_d] \in \mathbb{R}^{N\times d}$ corresponding to the first $d$ smallest eigenvalues (excluding the zero eigenvalue) using Eq. (\ref{eq:geigen}). The $i$-th region can now be represented as a $d$-dimensional vector $S_i^{\hat{Y}}$. The $k$-means algorithm is applied to group regions into $L$ segments, $\mathcal{Z} = \{\mathcal{Z}_k\}_{k=1}^{L}$, to produce a hard partition \cite{kim2013learning,shi2000normalized}. 
To obtain more reliable multi-class segmentation that can be generalized to arbitrary images with different number of classes, we treat it as a prior segmentation to provide the class likelihood for the multi-class segmentation. Note that our method can deal with the number of segments regardless of pre-defined $L$. We will discuss this in the experiments.

\begin{figure}[t]
	\begin{center}
		\includegraphics[width=0.9999\linewidth]{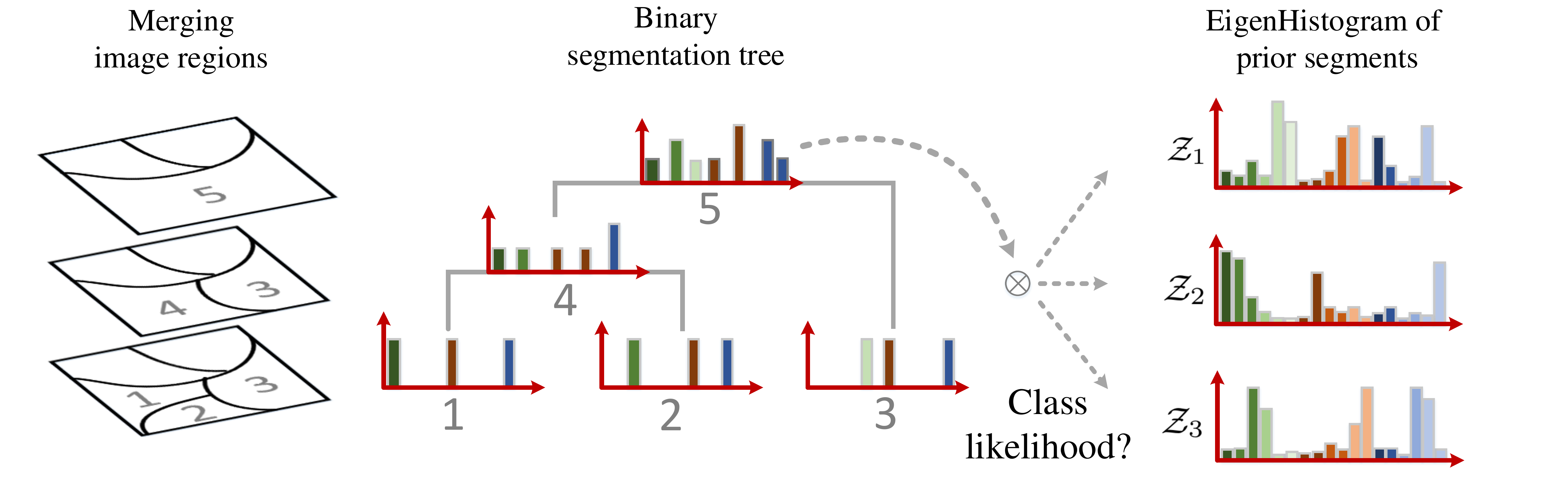}
	\end{center}

	\caption{In the merged binary segmentation tree, each leaf node representing an initial image region has an EigenHistogram while the EigenHistograms of the inner nodes are accumulated and normalized from that of their descendant nodes. Each node computes its class likelihood based on the EigenHistograms of the prior segmentation $\mathcal{Z}$. } \label{fig:eigenhis}
\end{figure}

For each dimension of $S_i^{\hat{Y}}$, we compute a histogram with $L$ bins uniformly spaced between $[0,1]$ based on the corresponding normalized eigenvector. As a consequence, a region will be represented as a $(d{\times}L)$-dimensional concatenated histogram (we set $d{=}6$ empirically and we will discuss the parameter $L$ in the experimental section). For each segment $\mathcal{Z}_k$, we accumulate and normalize the histograms of all regions belonging to this segment. We term this region representation as EigenHistogram (see Figure \ref{fig:eigenhis}).

\subsection{Multi-class segmentation}

Considering the image regions as a random field, we are interested in incorporating the unary potential (the class likelihood) of each region based on the prior segmentation $\mathcal{Z}$ and pairwise potentials between neighboring regions into a unified energy function, to achieve a holistic multi-class segmentation. Numerous literatures have investigated to learn effective unary potentials for random field based algorithms via structured support vector machine \cite{yang2013exemplar,lempitsky2011pylon} or convolution neutral network \cite{farabet2013learning,liu2015crf} to perform semantic segmentation. In contrast, EigenHistogram can be treated as a high-level representation which possesses spatial consistency, thereby intrinsically scalable to image segments of arbitrary size. Furthermore, it is easy and fast to compute without any supervision as other methods \cite{yang2013exemplar,lempitsky2011pylon,farabet2013learning} conduct. 

Following the Pylon model \cite{lempitsky2011pylon}, we can configure the regions into a hierarchical binary segmentation tree. Different from the traditional ``flat" random field models \cite{kolmogorov2004energy,boykov2001fast}, each node in our tree structure stands for a region nested from bottom to top, which enables the features to be extracted at different levels of the hierarchy to enrich the feature representation of the segments. In total, the constructed tree has $2N{+}1$ regions (the root node is the whole image), $\mathcal{S^{+}} = \{S_i|i = 1,...,2N{+}1\}$, which define a hierarchical random field.

Our goal is to assign labels $\bm{p}=[p_1,...,p_{(2{\cdot}N{+}1)}]^T$ to all regions in $S^{+}$. Therefore, we minimize the following object function:
\begin{equation}
\label{eq:crf}
\Theta(\bm p) = \sum_{i=1}^{2{\cdot}N{+}1} U(p_i) + \sum_{(i,j) \in \mathcal{N}(\mathcal{S})} B(S_i, S_j), \quad p_i \in\{0,1,...,L\},
\end{equation}
where $U(p_i)$ is the unary potential of the region $S_i \in \mathcal{S^{+}}$ to specify the cost of assigning label $p_i$ to $S_i$, and $B(S_i, S_j)$ is the pairwise potential to specify the boundary cost (exponentiated boundary strength \cite{yang2013exemplar}) between any two neighboring regions $(i,j) \in \mathcal{N}(\mathcal{S})$ in the child nodes, which is used to encourage the spatial smoothness. Note that $p_i$ is allowed to take a zero label such that it satisfies the non-overlapping requirement \cite{lempitsky2011pylon} by using the constraint:
\begin{equation}
\forall i \neq j, S_i, S_j \in \mathcal{S}^{+}, \; \text{if} \; S_i \cap S_j \neq  \emptyset, \; \text{then} \;  p_i \cdot p_j = 0,
\end{equation}
which ensures that any subtree can have only one single non-zero label.

Since we have clustered image regions into $L$ segments, the unary potential of region $S_i$ assigned to the $k$-th segment has the cost:
\begin{equation}
\label{eq:unary}
U_{p_i=k} = {-}\beta \cdot \langle \mathcal{H}(S_i),  \mathcal{H}(\mathcal{Z}_k)\rangle, \; S_i \in \mathcal{S}^{+}, \, \mathcal{Z}_k \in \mathcal{Z},
\end{equation}
where $U_{p_i}$ is the unary potential of the region $S_i \in \mathcal{S^{+}}$ to specify the cost of assigning label $p_i$ to $S_i$. $\beta$ determines the weight of the unary potential against the pairwise potential. $\mathcal{H}$ transforms a region into the EigenHistogram representation, where the class likelihood is calculated for each region in the tree. Following \cite{lempitsky2011pylon}, we compute the pairwise potentials as the exponentiated boundary strength.

EigenHistograms of the internal nodes of the binary segmentation tree are accumulated and normalized from that of the corresponding descendant nodes (see Figure \ref{fig:eigenhis}). Therefore, the partial non-smoothness effects of the eigenvectors (i.e., isolated regions as visualized in the right panel of Figure \ref{fig:eigventors}) reflected in the EigenHistograms of top-level nodes will be suppressed. Finally, we can leave the rest computation to the whole inference procedure to produce a holistic multi-class segmentation as the final output, by using the alpha-expansion based graph cut \cite{boykov2001fast}.

\begin{figure}[t]
	\begin{center}
		\includegraphics[width=.99\textwidth]{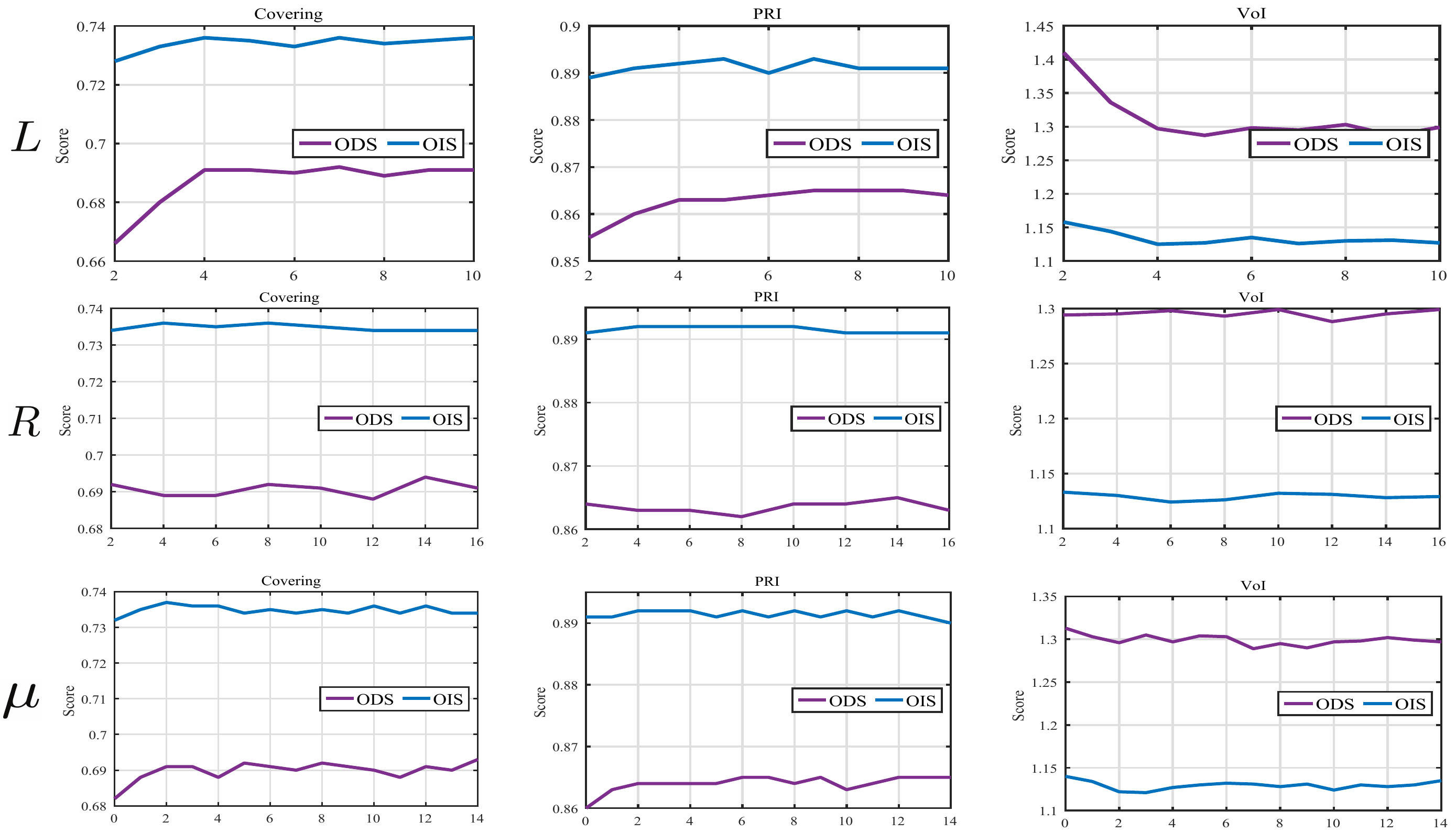}
		
	\end{center}
	\caption{Segmentation evaluation with different parameters on the training dataset. The top, middle, and bottom rows show the segmentation results by varying $L$, $R$, and $\mu$, respectively, as indicated in $x$-axis. It can be observed that the proposed method is insensitive to variations of the parameters.} \label{fig:parm}
\end{figure}

\begin{table}[t]
	\caption{The comparison of segmentation results and runtime on the BSDS500 dataset. }\label{table:bsds-seg} 

	\newcolumntype{C}{>{\centering\arraybackslash}p{3.0ex}}
	\centering	
	\begin{center} 
		\begin{tabularx}{.8\textwidth}{l|CC|CC|CC|c} 
			\hline
			\multirow{2}{*}{Method}	& \multicolumn{2}{c|}{Covering} &  \multicolumn{2}{c|}{PRI} & \multicolumn{2}{c|}{VoI} & \multirow{2}{*}{Time(s)}\\ \cline{2-7}
			& ODS   & OIS  & ODS  & OIS  & ODS   & OIS       \\
			\hline
			NCut \cite{shi2000normalized} & .45 & .53  &.78 &.80 & 2.23 & 1.89 & - \\
			Felz-Hutt \cite{felzenszwalb2004efficient} & .52 &.57 &.80 &.82 & 2.21 & 1.87 & - \\
			Mean Shift \cite{comaniciu2002mean} & .54 & .58  & .79 & .81 & 1.85 & 1.64 & - \\
			
			ISCRA \cite{ren2013image} & .59 & .66  & .82 & .85 & 1.60 & 1.42 & \textbf{$30$}\\
			gPb-owt-ucm \cite{arbelaez2011contour}   & .59  & .65     & .83 & .85  &  1.69 &  1.48 & \textbf{$240$}\\
			cPb-owt-ucm \cite{kim2013learning}   & .59  & .65     & .83 & .86  &  1.65 &  1.45 & \textbf{${>}240$}  \\ \hline
			red-spectral \cite{taylor2013towards} & .56  & .62    & .83 & .85    & 1.78 &   1.56 & ${\sim}12$ \\
			DC-Seg \cite{donoser2014discrete}  & .58   & .63       & .82     & .85   &  1.75 &   1.59  & \textbf{$6$}   \\
			$\text{DC-Seg}_{\text{full}}$ \cite{donoser2014discrete}  & .59   & .64      & .82     & .85   &  1.68 &   1.54  & \textbf{$144$} \\
			\hline 
			$\text{PMI}_{\text{low}}$ \cite{isola2014crisp} & .61 & .66  &.83	 &.86 & 1.58 & 1.42 & 30$^\star$ \\
			MCG \cite{arbelaez2014multiscale}  & .61   & .66      & .83     & .86   &  1.57 &   1.39  &  $18$ \\ 
			PFE$+$ucm \cite{yu2015piecewise} & .61 & .66  &.83 & .86 & 1.64 & 1.46 & ${>}900{\cdot}b^\dagger$ \\
			PFE$+$MCG \cite{yu2015piecewise} & .62 & .68  &.84 & .87 &1.56 & 1.36 & ${>}900{\cdot}b^\dagger$ \\ \hline
			Ours    & .62 & .66     & .83  & .86   & 1.59 &  1.43 & \textbf{$9$} \\ \hline
		\end{tabularx}	
		\begin{tablenotes}
			\item $^\star$Time is tested on half-sized images. $^\dagger b{=}\{4,8,16\}$ is the number of the embedding needs to compute.
		\end{tablenotes}

	\end{center}
\end{table}
\begin{figure}[!h]
	\begin{center}
		\includegraphics[width=0.99\linewidth]{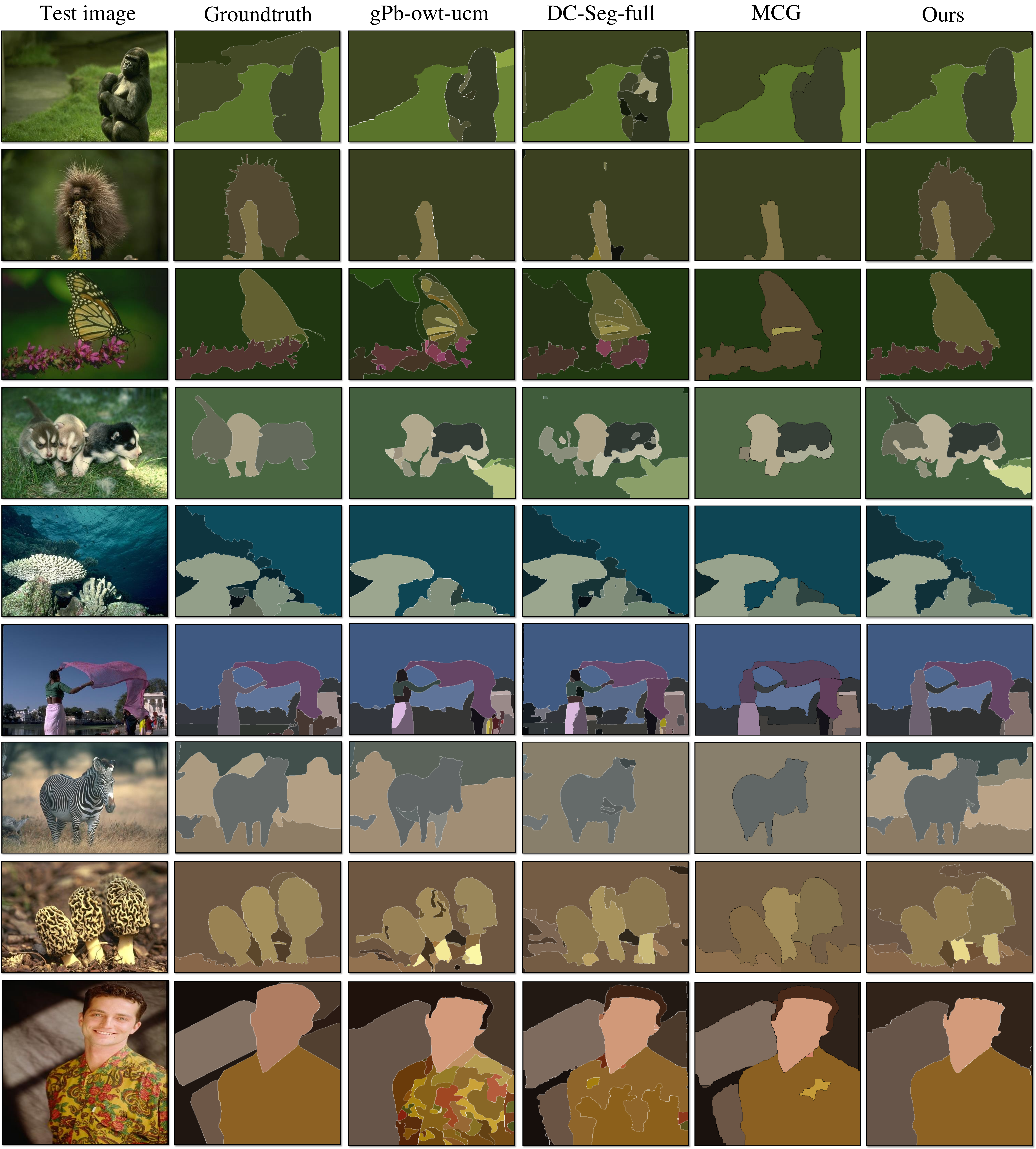}
	\end{center}
	\caption{Segmentation results on BSDS500. The first and second columns show the test images and the ground truth, respectively. The third to the sixth columns show the results obtained by gPb-owt-ucm \cite{arbelaez2011contour}, DC-Seg$_\text{full}$ \cite{donoser2014discrete}, MCG \cite{arbelaez2014multiscale} and our method, respectively. All results are visualized with the optimal scale (ODS) of the corresponding methods used for quantitative evaluation. Figure \ref{fig:partations} presents one graph partition result of the proposed method. } \label{fig:comare_images}
\end{figure}
\begin{figure}[!h]
	\begin{center}
		\includegraphics[width=0.9999\linewidth]{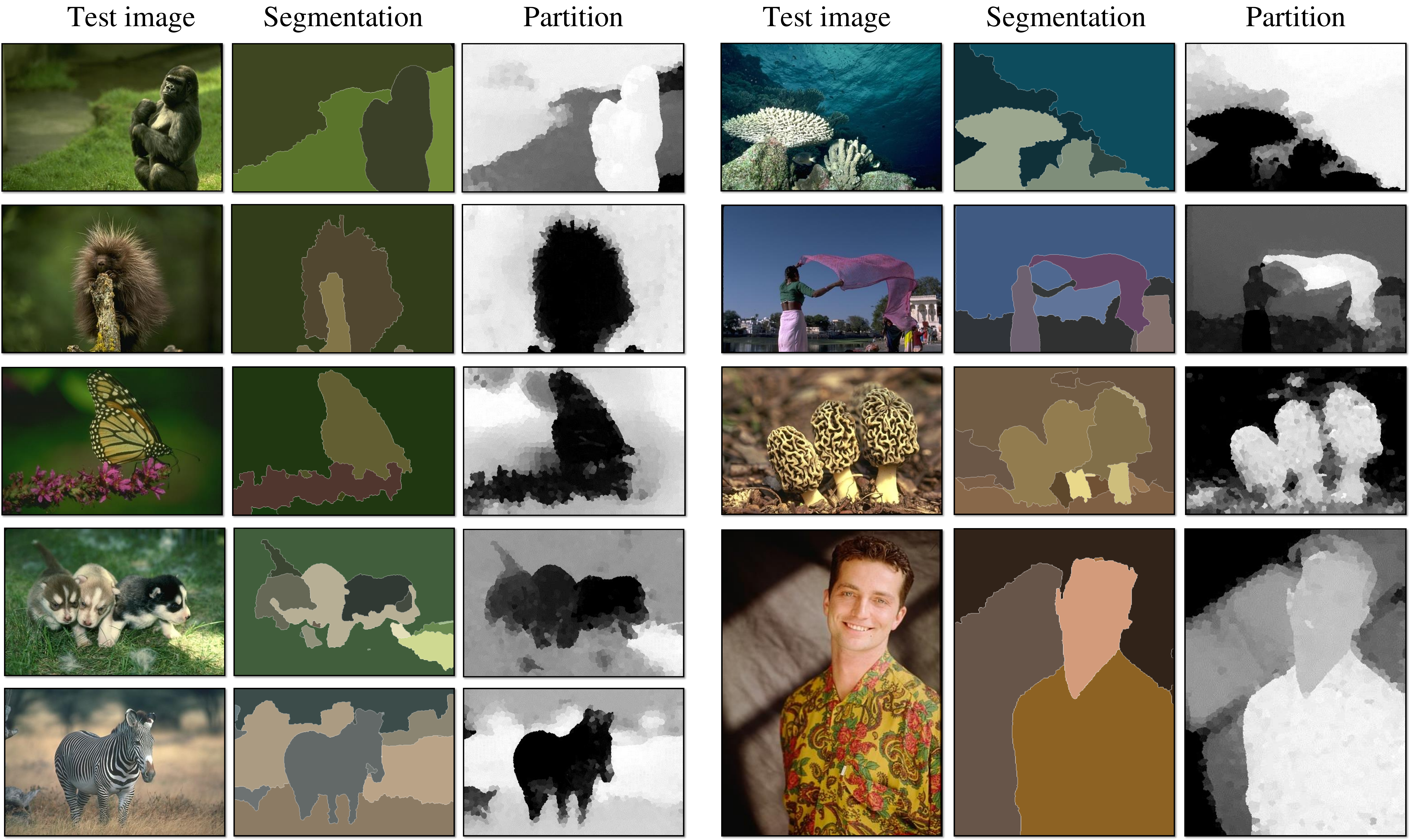}
	\end{center}
	\caption{The graph partition results of the proposed method on the test images in Figure \ref{fig:comare_images}. As can be observed, our graph partition is able to segment the objects from the background on the challenging images, where the objects have similar color or texture with that of background. } \label{fig:partations}
\end{figure}

\section{Experimental Results}
This section evaluates the segmentation performance of the proposed method. We first analyze the parameter settings. Then we evaluate and demonstrate the segmentation results, and compare to several state-of-the-art methods. 

We mainly evaluate the proposed segmentation approach using the challenging Berkeley Segmentation Dataset (BSDS500) \cite{arbelaez2011contour}. BSDS500 is widely used as the benchmark for image segmentation and boundary detection, which contains 200 training, 100 validation, and 200 test images. We use several standard evaluation criteria \cite{arbelaez2011contour} to conduct quantitative analysis: 
Segmentation Covering, Probability Rand Index (PRI), and Variation of Information (VoI), which measure per-pixel segment overlapping, pairwise pixel matching, and segmentation-wise entropy, respectively. For each measurement, we report the values with the optimal dataset scale (ODS) and optimal image scale (OIS). 
We further evaluate our method on large-scale PASCAL VOC and COCO datasets to show the generalization ability of our method for object segmentation and compare to two state-of-the-art methods.

\subsection{Implementation details}
We investigate the parameter sensitivity of the proposed method and select the optimal values based on the training set. Then we apply these values to the independent test set.

Figure \ref{fig:parm} shows the performance of the proposed method with respect to $L$ for clustering, $R$ for selecting nearest regions in Eq. (\ref{eq:Kmat}),  and $\mu$ for graph partitioning. The selection of the optimal value of the number of segments $L$ is dependent on the test set, but we do not select the best $L$ based on the test set, by which we aim to demonstrate the strong generalization ability of the proposed method.
For the parameter $\mu$, compared with $\mu=0$, which means that only $E_{local}$ is considered in the energy function, $E_{global}$ with $\mu=8$ improves the accuracy by ${>}.1$ODS (Covering). Section 5.4.2 further validates the effectiveness of $E_{global}$.
As can be observed, the proposed method is insensitive to the three parameters. As a result, we set $L=6$, $\mu=8$, and $R=14$ throughout the following experiments.

We empirically set $e_1=20$ for kernel density estimation in Eq. (\ref{eq:cooccur}) and $e_2=40$ for computing $W$ in Eq. (\ref{eq:affmat}). Since the test set has approximately equal image sizes, we can assume that these two values can be generalized to all test images. We empirically found that the parameter $\beta$ in Eq. (\ref{eq:unary}) varies from images to images. In practice, we run the inference procedure to obtain multiple segmentations by varying the $\beta$ value between $[200,300,...,800]$ with an interval equal to $100$, and take the average of all these outputs and the superpixel map as the final segmentation.

We use the toolbox provided by Doll{\'a}r \textit{et al.} \cite{dollar2015pami} to generate the superpixel map (i.e., the structured edge (SE) detector followed by UCM) with roughly uniform region sizes. Our implementation is based on Matlab running on a standard Intel i7 desktop.

\begin{table}[t]
	\caption{The running time (in second) of each phase of the proposed method.  }\label{table:timecost} 
	\newcolumntype{Y}{>{\centering\arraybackslash}X}
	\newcolumntype{C}{>{\centering\arraybackslash}p{3.9ex}}
	\small
	\begin{center}
		\begin{tabularx}{.72\textwidth}{l|c|c|c|c}
			\hline
			Phase                                                & Min & Max   & Mean & Var. \\
			\hline
			1: Region structure generation & 2.60 & 3.73  & 3.08 & 0.05 \\ \hline
			2: Graph construct. and partition                                & 3.30 & 6.89  & 4.51 & 0.56 \\ \hline
			3: Multi-class segmentation                          & 1.39 & 2.26  & 1.71 & 0.03 \\ \hline
			Total                                                & 7.54 & 12.6 & 9.30 & 1.00 \\ \hline
		\end{tabularx}
		
	\end{center} 
\end{table}

\subsection{Segmentation result comparison}
We evaluate the performance and efficiency of the proposed method, and compare it to several state-of-the-art methods.	

In Table \ref{table:bsds-seg}, we compare the proposed approach to several state-of-the-art methods in terms of segmentation accuracy and running time on the BSDS500 test set. As one can see, the proposed method significantly outperforms most of the comparative methods. $\text{PMI}_{\text{low}}$ \cite{isola2014crisp} is a boundary detection method, which embeds the edge map into OWT-UCM \cite{arbelaez2011contour} to obtain accurate segmentation. We report its the best accuracy, which is achieved on low resolution images. The recently proposed multiscale combinatorial grouping (MCG) \cite{arbelaez2014multiscale} and piecewise flat embedding (PFE) \cite{yu2015piecewise} obtain significant improvement compared with the early method, such as red-spectral \cite{taylor2013towards} and DC-Seg \cite{donoser2014discrete} (see Table \ref{table:bsds-seg}). MCG uses hierarchical UCMs to boost the segmentation performance. PFE integrates its computed graph partitions into the gPb-owt-ucm \cite{arbelaez2011contour} and MCG, which achieves good segmentation performance. However, PFE suffers from the computationally expensive optimization. The proposed method outperforms PEF$+$owt-ucm and it achieves close segmentation performance compared with PFE$+$MCG. More importantly, the proposed method is hundreds of times faster than the PFE based methods. DC \cite{donoser2014discrete} and red-spectral \cite{taylor2013towards} also emphasize on fast image segmentation, but their segmentation accuracy is not as accurate as ours.

\begin{figure}[t]
	\begin{center}
		
		\includegraphics[width=0.999\linewidth]{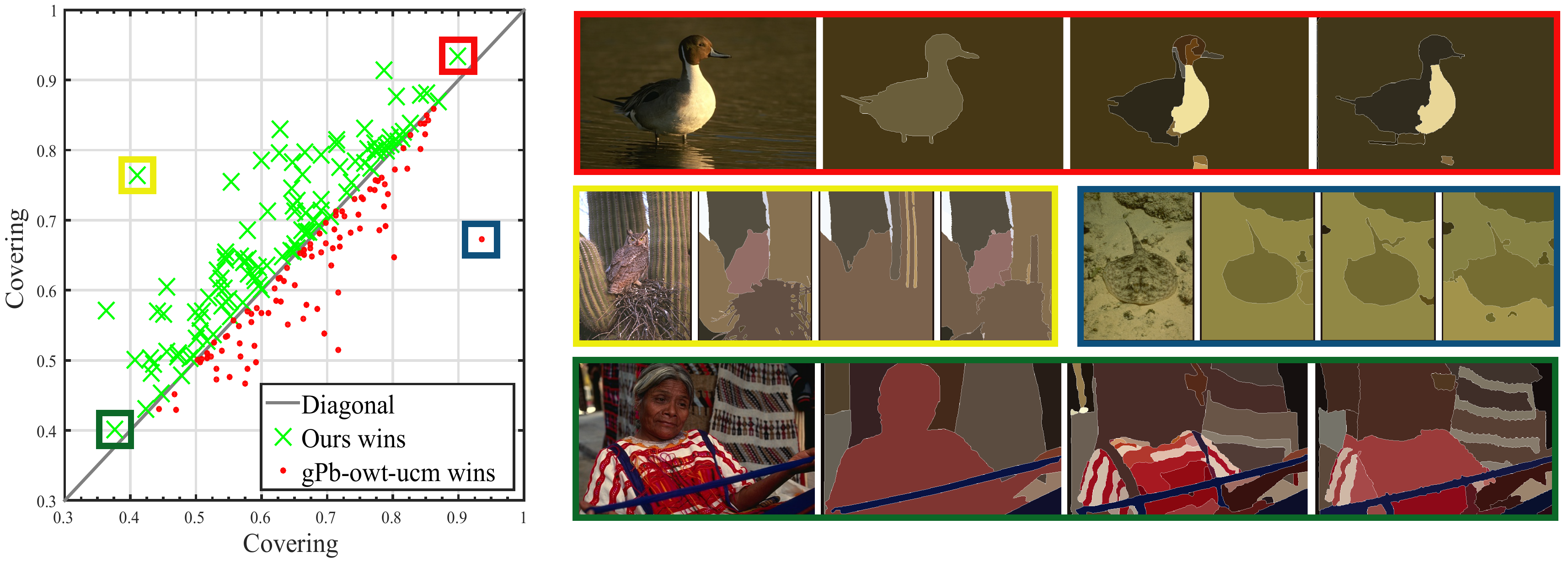}
		
	\end{center}
	\caption{Pairwise segmentation comparison between gPb-owt-ucm \cite{arbelaez2011contour} and our method on the $200$ test images. In the left figure, $\textcolor{green}{\times}$ above the diagonal indicates the image our method wins, and $\textcolor{red}{\bullet}$ below the diagonal indicates gPb-owt-ucm wins (best viewed in electronic form). We also highlight several extreme cases corresponding to two with the largest discrepancy and the worst and the best cases. Each is marked with an unique colored rectangular. In each rectangular of the right slide containing four images, from left to right, shows the test image, ground truth, the gPb-owt-ucm result, and our result, respectively.} \label{fig:pairwise-comparison}
\end{figure}

Figure \ref{fig:comare_images} shows the qualitative segmentation results. Figure \ref{fig:partations} presents the graph partitioning result obtained by the proposed method, which provides good initial segmentation proposals. Compared with other methods, the proposed method is able to resist the object internal variances to avoid small segments, so that the segments are much more spatially consistent.  In addition, the proposed method can implicitly figure out the best number of segments regardless of the pre-defined $L$ value. It is because that
EigenHistogram can penalize over-segmentation since homogeneous segments have similar EigenHistogram and thus proximate unary potentials, encouraging them to be merged. The first three rows of Figure \ref{fig:comare_images} particularly highlight the above-mentioned capability. 
To provide more detailed comparison, in Figure \ref{fig:pairwise-comparison}, we show the pairwise segmentation results obtained by our proposed method compared to the classical gPb-owt-ucm \cite{arbelaez2011contour} method. As can be observed, the proposed method shows obvious improvement on a large number test images.

\vspace{.1cm}
\noindent
\textbf{Time Efficiency:} 
The comparison of running time is shown in the rightmost column of Table \ref{table:bsds-seg}. The test image size is $321\times 481$. The proposed method is much faster than other competing methods because of several important aspects:

\begin{enumerate}
	\itemsep-.1em
	\item The proposed method does not need complex feature computation, which is superior than gPb based methods \cite{yu2015piecewise,arbelaez2011contour}.
	\item  We construct the graph model based on superpixels rather than raw pixels. Although we incorporate multiple cues into a graph with complicated constraints, the graph partitioning is a single eigenvector system. While in the PFE method \cite{yu2015piecewise}, performing graph partation is particularly computationally expensive.
	\item EigenHistogram is efficient to compute and very scalable to regions with arbitrary size for hierarchical multi-class segmentation.
\end{enumerate}
The proposed method is executed in three phases: 1) generating the superpixel map and constructing the hierarchical segmentation tree, 2) constructing and partitioning the graph, and 3) conducting multi-class segmentation. Given an $H\times$W resolution image, phase 1 takes low logarithmic time of random forest tree depth to predict edge map with a random forest, $O(HW+N)$ to compute superpixels with $N$ regions, and $(\log N)$ to construct the hierarchical binary tree. Phase 2 takes up to a factor of $O(HW {+}N)$ to compute all image features with respect to pixels and regions and approximately $O(fN^2)$ to compute the affinity matrix, where $f$ is the feature dimension. Since $(D-W+\mu M)$ in Eq. (\ref{eq:geigen}) is sparse, solving the eigen decomposition problem with a $N\times N$ affinity matrix takes $O(N(\tilde{R}{+}R))$ using a Lanczos algorithm according to \cite{shi2000normalized}, where $\tilde{R}{+}R {\ll} N$ is the adjacencies from both local ($\tilde{R}$) and global connections ($R$) in the graph. 
Phase 3 optimizes Eq. (\ref{eq:crf}) with $L$ classes with approximately $O(N^2L)$ using graph cut with alpha-expansion \cite{kolmogorov2004energy,lempitsky2011pylon}. Therefore, the overall method has approximate time complexity $O(HW{+}fN^2)$, bounded by phase 2.

We also evaluate the detailed running time of the proposed method on the $200$ BSDS500 test  images. Table \ref{table:timecost} shows the detailed time cost of each phase. Compared with most comparative methods, the proposed method is more scalable for practical usages.

\begin{figure*}[!t]
	\begin{center}
		\begin{subfigure}[b]{0.49\textwidth} \includegraphics[width=\textwidth]{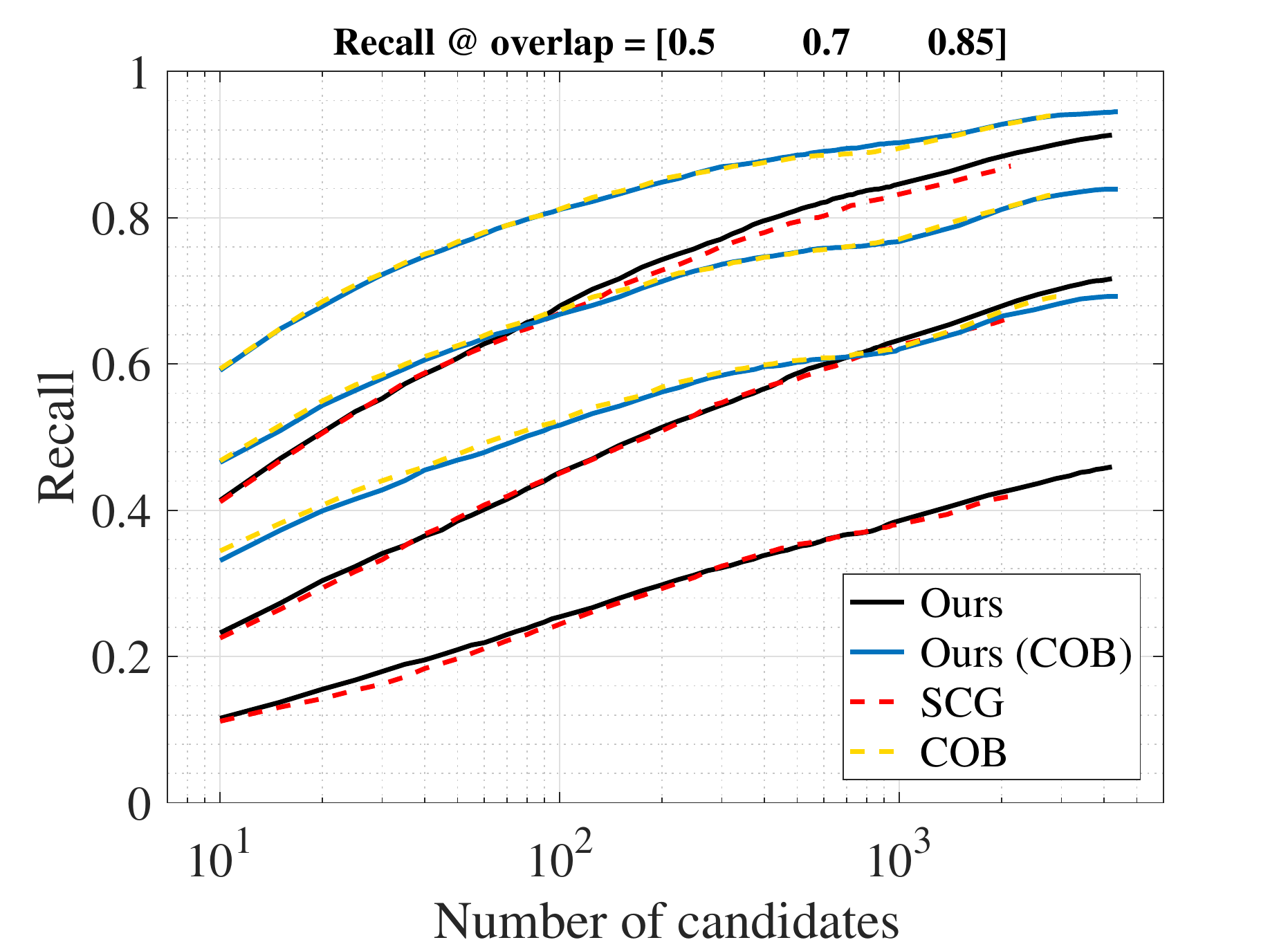}	\end{subfigure}
		\begin{subfigure}[b]{0.49\textwidth} \includegraphics[width=\textwidth]{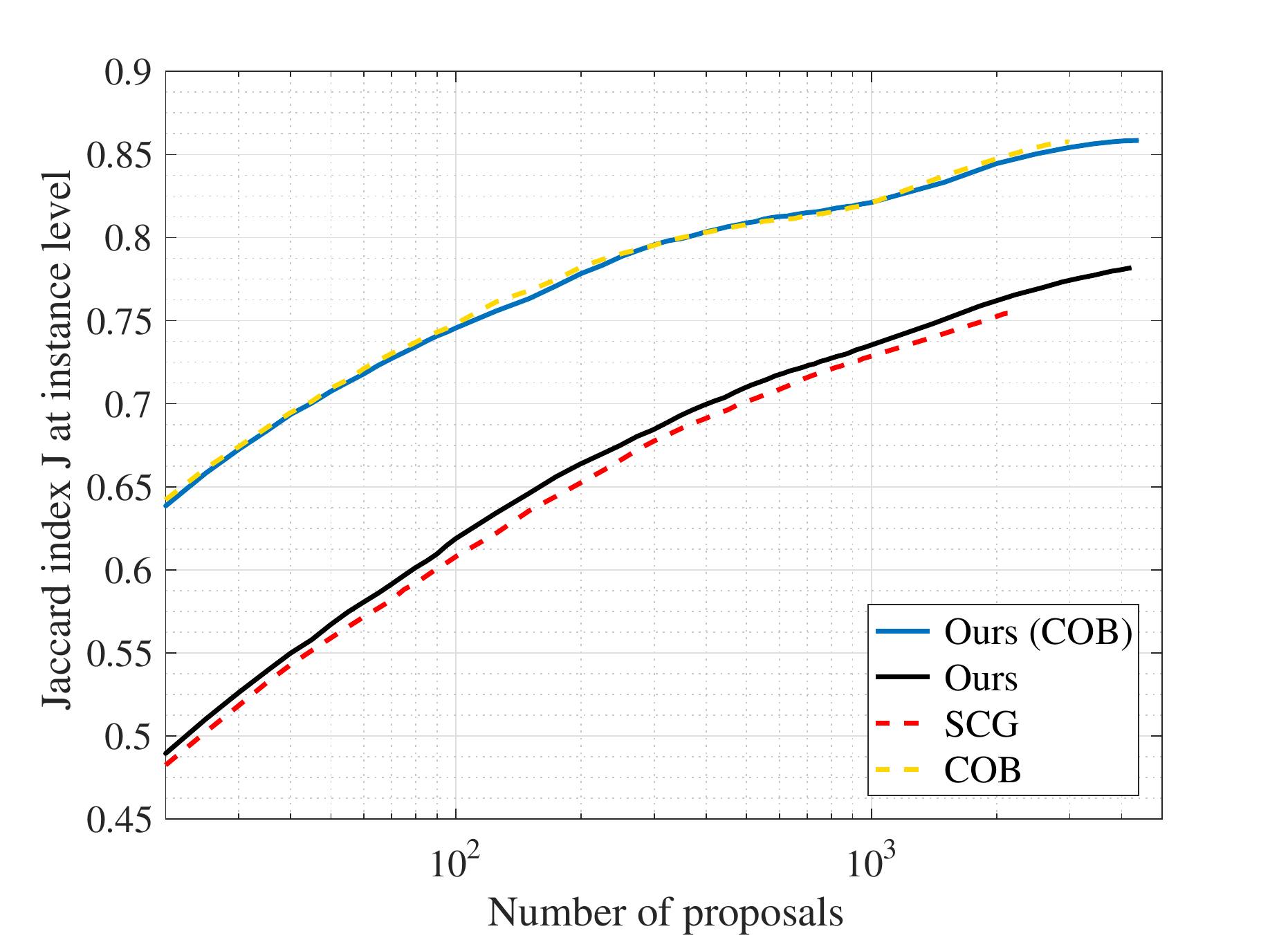}	\end{subfigure}
	\end{center}
	\vspace{-.5cm}
	\caption{Object segmentation evaluation on PASCAL VOC 2012. The different lines in the same color indicates results under different Jaccard overlapping thresholds. \cite{pont2017multiscale} introduces detailed evaluation metrics. } \label{fig:pascal}
	
\end{figure*}
\begin{figure*}[!t]
	\begin{center}
		\begin{subfigure}[b]{0.49\textwidth} \includegraphics[width=\textwidth]{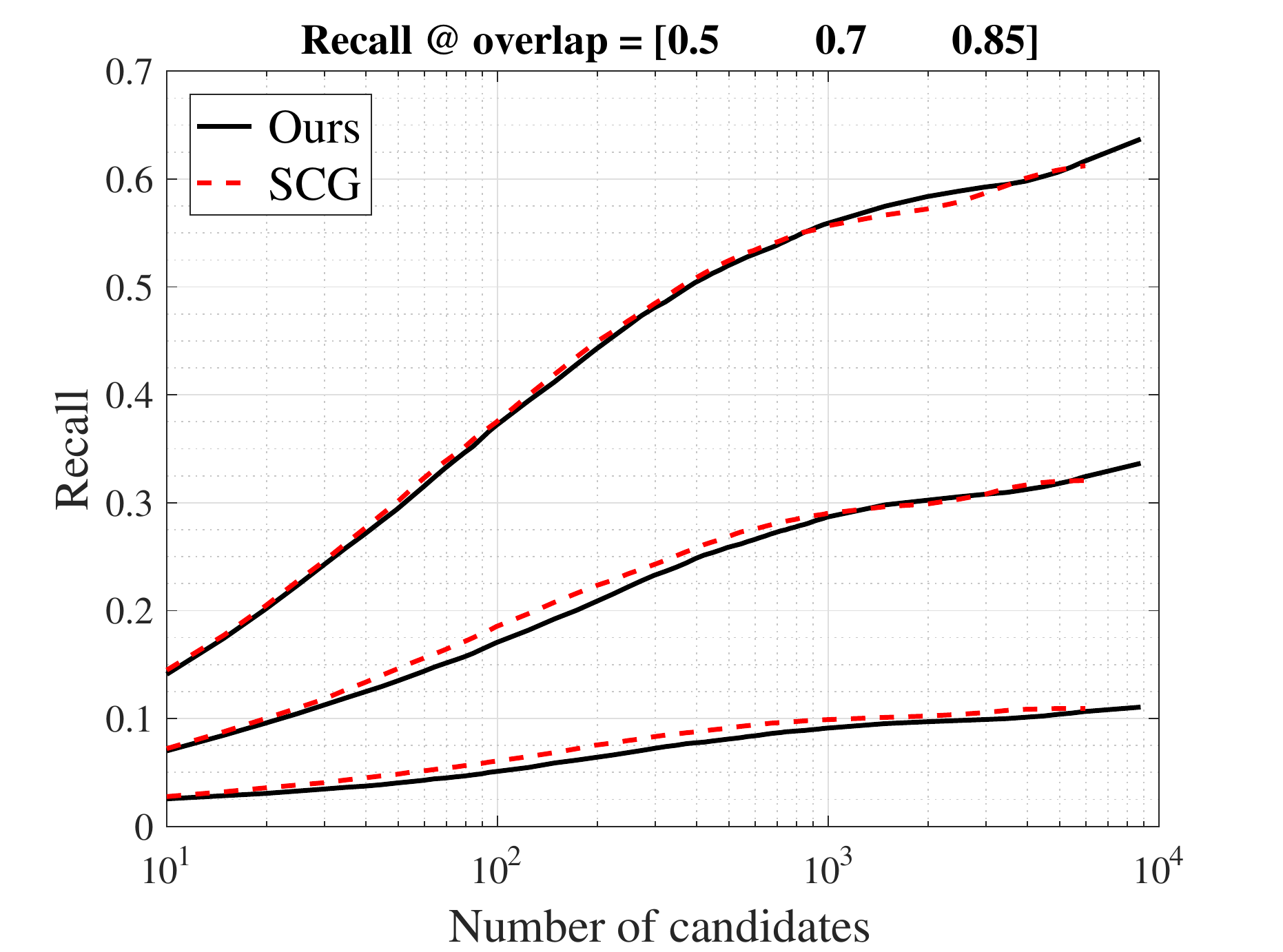}	\end{subfigure}
		\begin{subfigure}[b]{0.49\textwidth} \includegraphics[width=\textwidth]{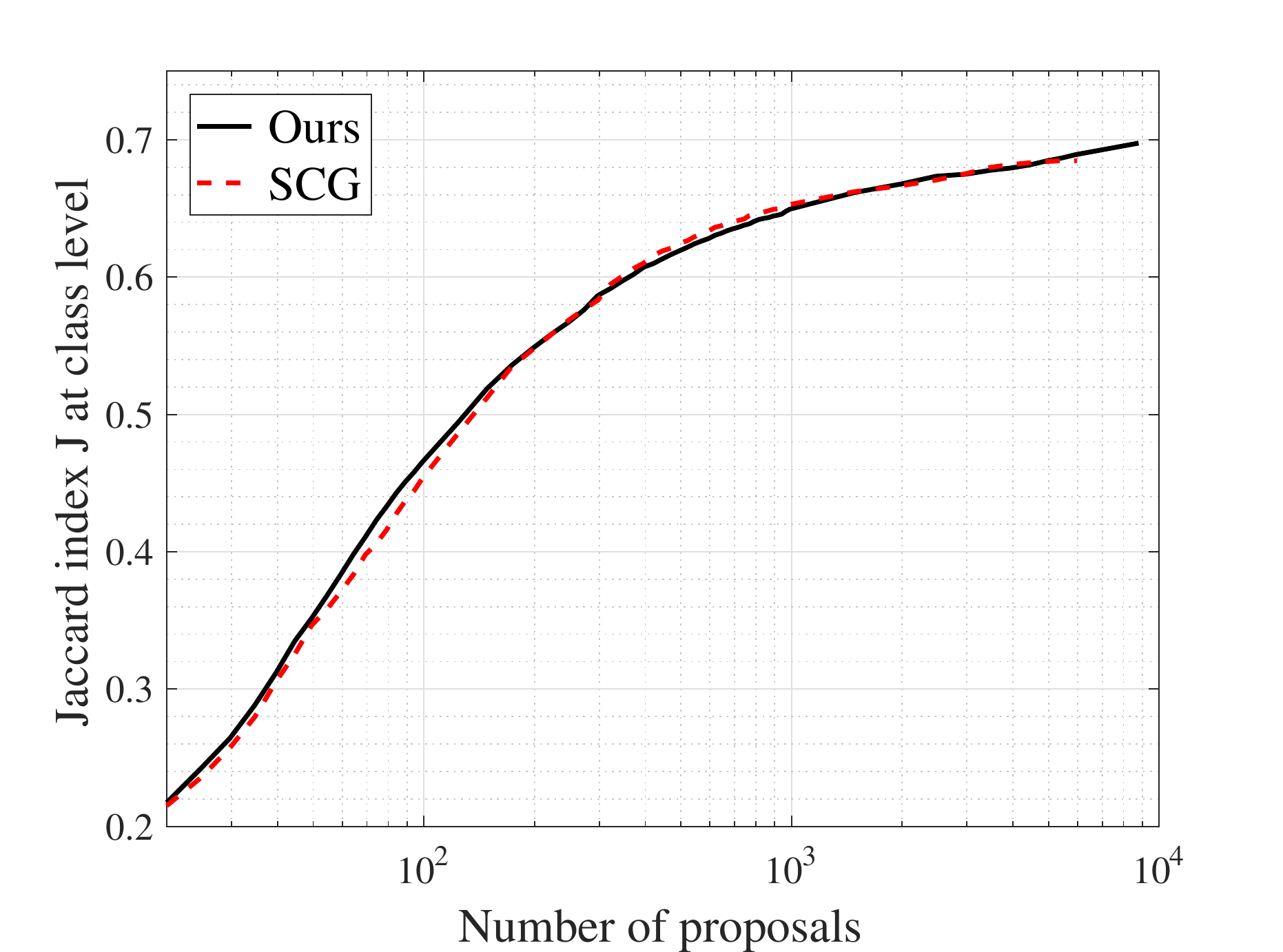}	\end{subfigure}
	\end{center}
	\vspace{-.5cm}
	\caption{Object segmentation evaluation on COCO. See text for explanations.} \label{fig:coco}
\end{figure*}

\begin{figure*}[!t]
	\begin{center}
	\includegraphics[width=\textwidth]{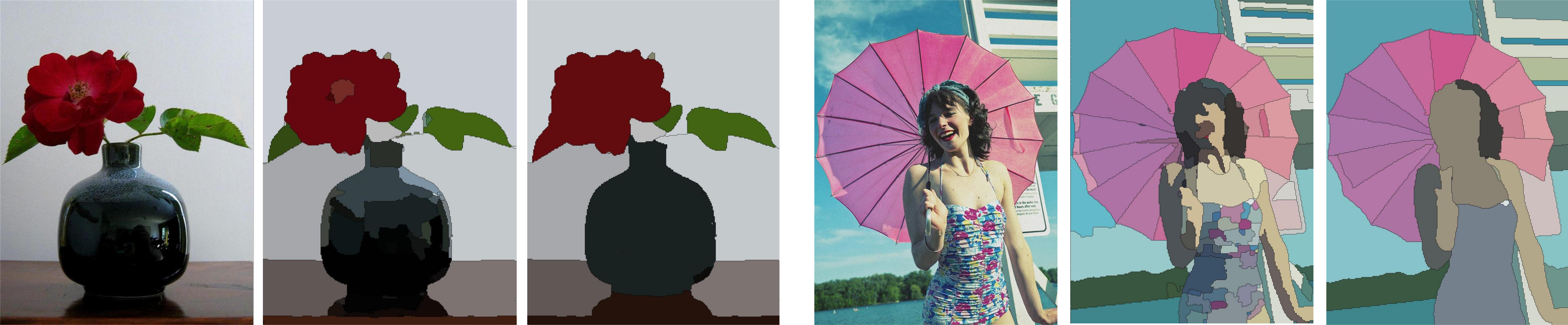}	

	\end{center}
	\vspace{-.5cm}
	\caption{Qualitative evaluation of the segmentation on COCO. Our method (right column) generates obviously clearer segmentation results with significantly reduced over-segmentation (e.g. the flower and vase in the left images and swimwear in the right images) than the comparing method (middle column) \cite{pont2017multiscale}. } \label{fig:coco_compare}
\end{figure*}
\subsection{Towards large-scale object segmentation}
This section further demonstrates our proposed method on the large-scale PASCAL VOC \cite{everingham2010pascal} and COCO \cite{lin2014microsoft} segmentation datasets\footnote{According to the experiment settings of \cite{pont2017multiscale}, for PASCAL, 1,464 training images and 1,449 validation images are used. COCO totally contains 82,783 training images and 40,504 validation images in total. In our experiments, we randomly select 5,000 and 2,500 from the training and validation set, respectively for evaluation.}. Since our method generates region segmentation composed by a set of connected regions (the same as UCMs),
we can fully use our method to generate object proposals by training an object proposal grouping classifier following \cite{pont2017multiscale}. We closely follow its training procedure and evaluation settings. In brief, Jaccard Index $J$, i.e. the size of the intersection of the pixel union of two regions, is used to evaluate the accuracy of generated objects compared with groundtruth.

Figure \ref{fig:pascal} shows the comparing results for PASCAL VOC. We compare with a method proposed by \cite{pont2017multiscale}, denoted as singlescale combinatorial grouping (SCG).  As can be observed on the two evaluation metrics, our method improves the performance of SCG on the recall evaluation metrics consistently. We also compare with a recent deep learning based method COB \cite{Man+17} which aims at detecting accurate object boundaries. It combines with MCG \cite{pont2017multiscale} to perform object segmentation and achieved significant improvement. Note that region segmentation highly relies on the quality of boundary detection (it is out of the focus of this paper). As will demonstrated in Section 4.1, our method is flexible to be an extension of arbitrary baseline methods. Hence, we use the edge maps generated by COB (denoted as Ours (COB)). As can be observed, our method improves a substantial margin compared with our original method and achieves competitive results compared with COB.
Figure \ref{fig:coco} compares the results on COCO. Our method shows better results than SCG (right) at low numbers of proposals and competitive results on the recall with respect to the number of candidates.

SCG is designed for generate image object candidates, so its generated UCMs contain very fine and small region segments, which is an advantage when computing evaluation metrics for images with multiple objects. However, our method does not have designs for this goal. Compared with it, our method is significantly more proficient at segmenting the salient objects in images. We will further analyze this behavior in the next section. The PASCAL dataset is mainly collected for image and object segmentation tasks. According to our observation, PASCAL images usually contain definite and salient objects. Therefore, our method performs better and largely improves SCG.
While in COCO, most of images are outdoor scenes that usually contain many small and indefinite objects. That is the reason why the improvement on COCO for our method is not as large as that in PASCAL, compared with SCG. We qualitatively compare with SCG on COCO images with relatively definite objects. As can be seen in Figure \ref{fig:coco_compare}, our method can significantly reduce over-segmentation and give rises to clearer segmentation results.
Nevertheless, the shown results on the two large-scale datasets are sufficient to demonstrate the generalization ability of our proposed method to different datasets with diverse scenes \footnote{Note that we did not select the parameters of our method on the targeting training datasets following Section 5.1 but used the unique one selected using the BSDS500 training set. We believe there is still room for improvement with careful fine-tunning.}.    

\begin{table}[t]
	\caption{The segmentation results under different configurations. Different region generation baselines which are used by our method are indicated in $(\cdot)$. The last row is the result of our method (SE+ucm) when $E_{global}$ is not applied. Please see text for detailed explanations. }\label{table:bsds-baseline-compare}
	\newcolumntype{C}{>{\centering\arraybackslash}p{3.0ex}}
	\centering	
	
	\begin{center} 
		\begin{tabularx}{.785\textwidth}{l|CC|CC|CC} 
			
			\hline
			\multirow{2}{*}{Method} & \multicolumn{2}{c|}{Covering} &  \multicolumn{2}{c|}{PRI} & \multicolumn{2}{c}{VoI}  \\  \cline{2-7}
			& ODS   & OIS     & ODS  	  & OIS   & ODS    & OIS       \\ \hline
			
			SemiContour+ucm			&.56		&  .63		&	.82		&	.85	&	1.79	& 1.57	\\
			Ours (SemiContour+ucm)	& 	.60		&	.64		&	.83 &	.85	& 1.68	& 1.50	\\  \hline
			MCG 		  	& .61   & .66     & .83     & .86   &  1.57  &   1.39 \\
			Ours (MCG)    	& .62   & .66     & .84     & .86   & 1.57   &  1.40  \\ \hline
			SE+ucm     			 & .59   & .64     & .83     & .86	& 1.71		&  1.51 \\
			Ours (SE+ucm)    	& .62   & .66     & .83     & .86   & 1.59   &  1.43  \\ \hline
			
		\end{tabularx}	
		
	\end{center}
\end{table}
\begin{figure*}[t]
	
	\newcolumntype{C}{>{\centering\arraybackslash}p{3.0ex}}
	\centering	
	\begin{minipage}[b]{0.10\linewidth}
		\centering
			\begin{tabularx}{.4\textwidth}{|CCCC|CC|}
				\cline{1-6}
				\multicolumn{4}{|c|}{Components} & \multicolumn{2}{c|}{Covering}   \\ \cline{1-6}
				$E_l$  & $E_g$  & $MS$  & $kM$   & ODS & OIS  \\ \cline{1-6}
				\checkmark  &  &    & \checkmark             & .43 & .50  \\ \cline{1-6}
				\checkmark  & \checkmark  &    & \checkmark             & .52 & .52    \\ \cline{1-6}
				\checkmark  &  &  \checkmark    &            & .60 & .65   \\ \cline{1-6}
				\checkmark  & \checkmark & \checkmark    &        & .62 & .66   \\ \cline{1-6}
			\end{tabularx}	
		\vspace{.4cm}
	\end{minipage}\hfill
	\begin{minipage}[b]{0.5\linewidth}
			\centering
			\includegraphics[width=0.8\textwidth]{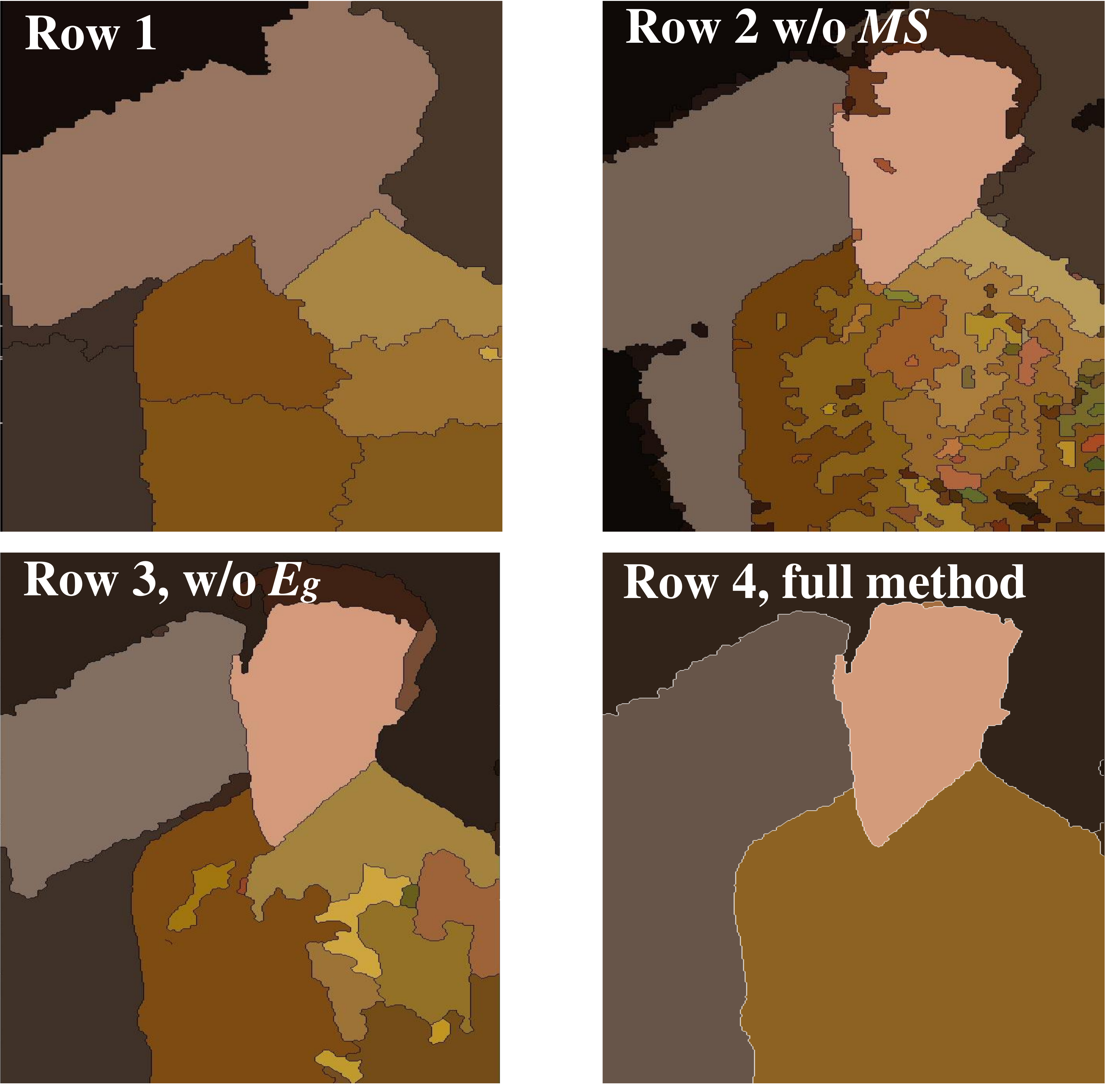}	
	\end{minipage}
\caption{Ablation study to analyze the effectiveness of each component of the proposed method. \textbf{Left}: Each row shows the result of a combination of the components. $E_l$, $E_g$, $MS$, and kM denote $E_{global}$, $E_{local}$, multi-class segmentation, and k-means, respectively. To evaluate the effectiveness of $MS$, we simply use k-means to cluster eivenvectors (11 classes), as an alternative to perform $MS$. The 4th row is our full method. \textbf{Right}: The qualitative results corresponds to each row of the left side table. As can be observed from both quantitative and qualitative results, the proposed $E_{g}$ and $MS$ components play important roles in generating clear segmentation and better scores. 
}\label{table:bsds-ablation}
\end{figure*}

\subsection{Analysis}

\subsubsection{Serving as an extension to improve baseline methods} 
We consider the cases of using different methods to generate superpixel maps as the input of the proposed method, which allow us to conduct more detailed analyses. It is necessary to notice that, although the proposed method is flexible to build upon these methods, it is not an extension of the underling methods. In contract, the proposed method is a new exploration of accurate and fast spectral clustering based image segmentation.
In addition, many state-of-the-art methods use accurate supervised edge detectors and other trained classifiers \cite{arbelaez2014multiscale,chenscale}. We are particularly interested in reducing number of training data with an aim to completely unsupervised image segmentation. Either unsupervised \cite{li2015unsupervised} or semi-supervised SE detector \cite{zhang2016semicontour} can be used as the underlying edge detectors.
We consider using the latter, namely SemiContour \cite{zhang2016semicontour} ($3$ training images are used), as an alternative to the originally used SE. We compare the performance in Table \ref{table:bsds-baseline-compare}. The obtained segmentation results consistently improve the segmentation accuracy of different baseline methods. Particularly, we observe $.4$ODS (Covering) improvement over SemiContour+ucm and $.3$ODS improvement over SE+ucm.
\begin{figure}[t]
	\begin{center}
		
		\includegraphics[width=0.99\linewidth]{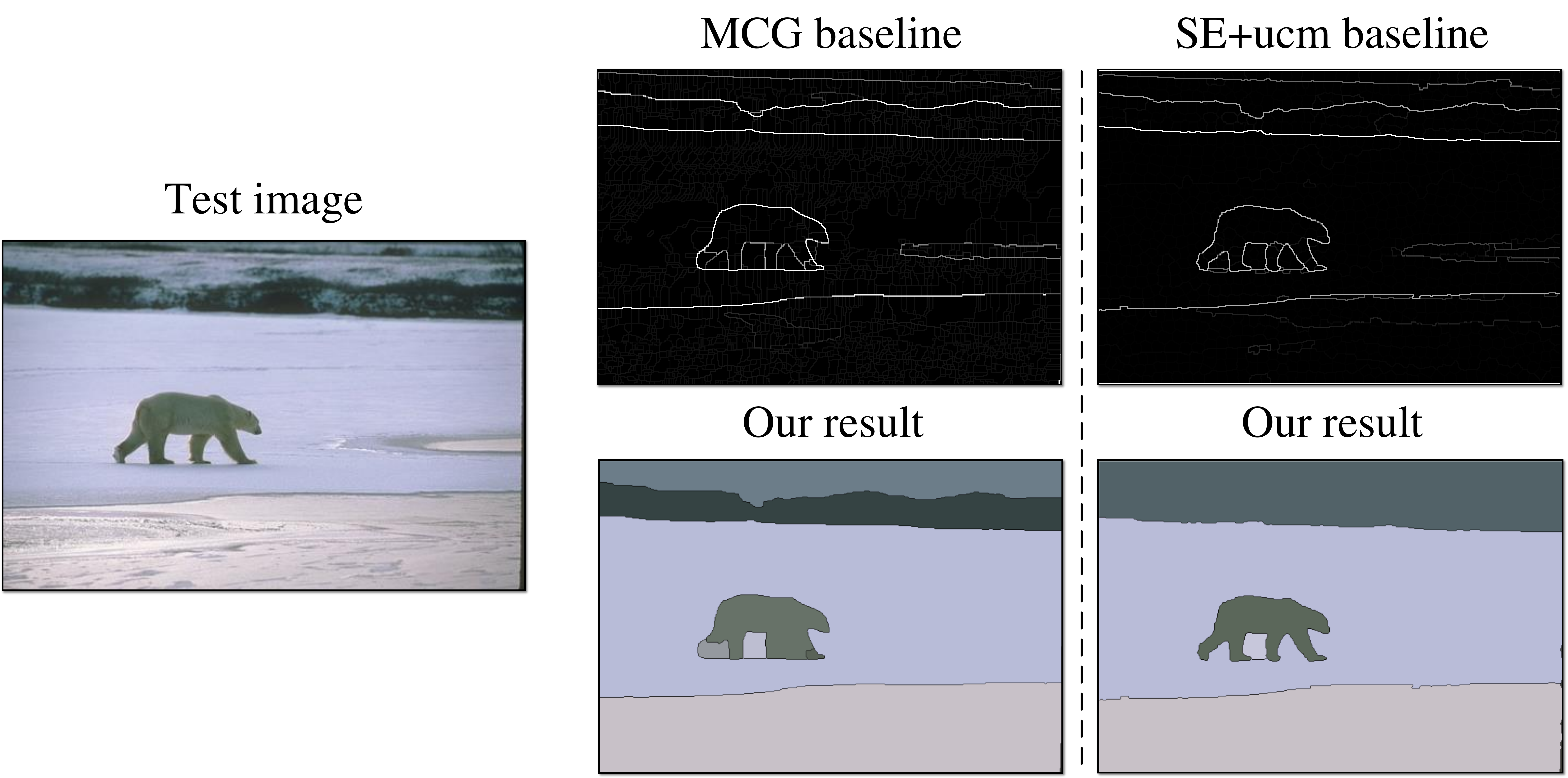}
		
	\end{center}
	
	\caption{The segmentation results (shown in the second row of right slide) of the proposed method which using MCG and SE+ucm to generate superpixel maps as candidate regions (the first row), respectively. Note that MCG sharpens noisy edges below the bear (upper left), resulting in worse segmentation (bottom left). } \label{fig:mcg_se_compare} 
\end{figure}
\begin{figure}[t]
	\begin{center}
		\includegraphics[width=0.6\linewidth]{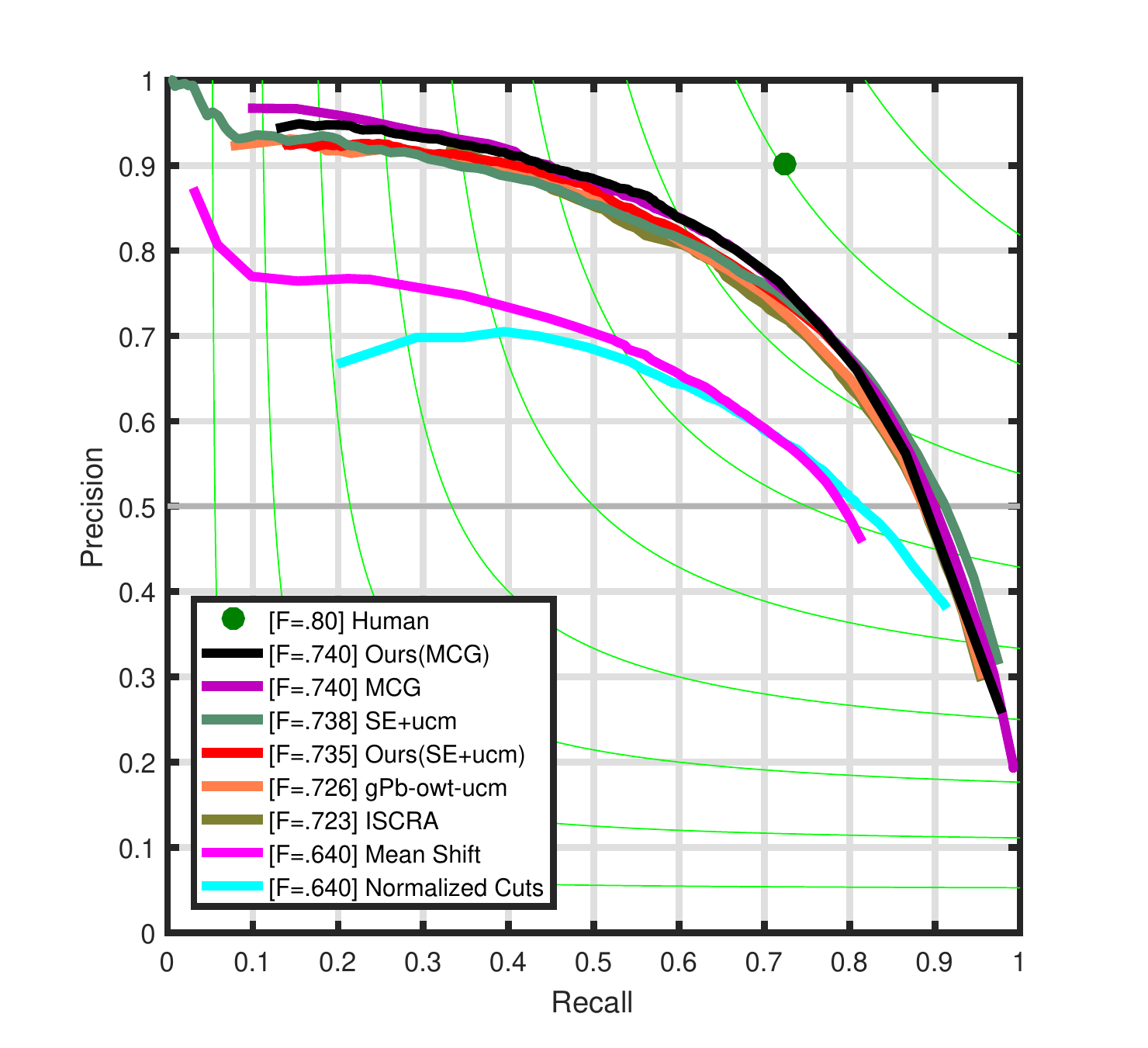}
	\end{center}
	
	\caption{The Precision-Recall curves obtained by the proposed method and competing methods for boundaries on the BSDS500 dataset. $(\cdot)$ indicates the baseline method used by the proposed method.}\label{fig:edges}
	
\end{figure}

\begin{figure}[t]
	\begin{center}
		
		\includegraphics[width=0.99\linewidth]{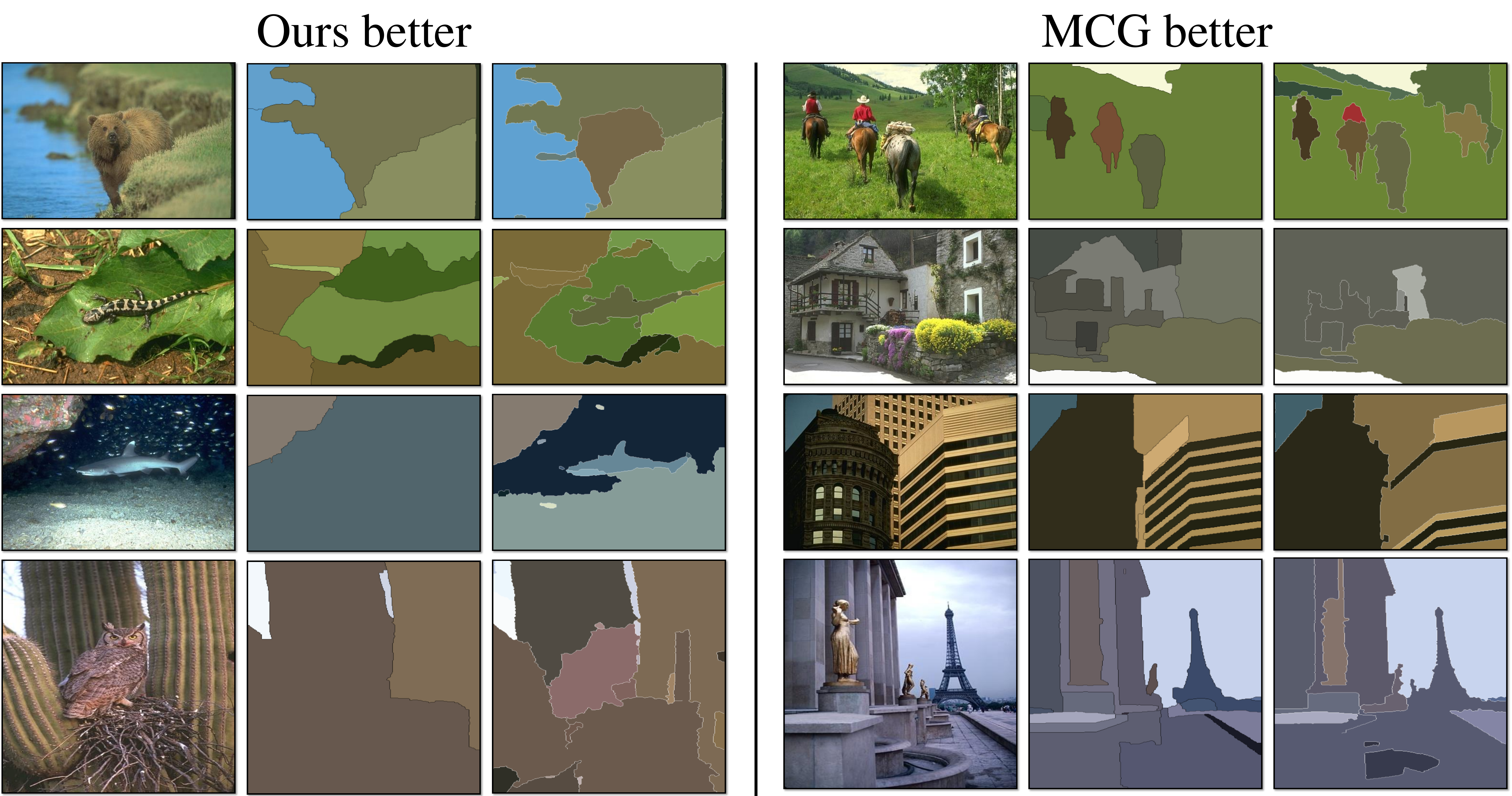}
		
	\end{center}
	
	\caption{Each sample shows the segmentation results of our proposed method with SE+ucm baseline (right) and MCG (middle) \cite{arbelaez2014multiscale}. The left side shows images that our method wins and the right side shows images that MCG wins.} \label{fig:comare_to_mcg}
	
\end{figure}

\subsubsection{Ablation study} 
We analyze the effectiveness of each component of the proposed method. The proposed global connection (Section 3.2) is very effective at capturing the affinity between spatially distant regions belonging to the same objects. And the proposed multi-class segmentation is critical to generate smooth and clear segmentation map and makes our method robust to arbitrary images. Figure \ref{table:bsds-ablation} evaluates each components both qualitatively and quantitatively. 
Comparing with our method without using $E_{global}$, we observe obvious improvement (comparing the 3rd row against 4th row and the 1st row against the 2nd row), which indicates the effectiveness of the proposed energy term $E_{global}$. To validate our multi-class segmentation, we conduct an experiment by simpling clustering the generated graph partitions (i.e. eigenvectors) using k-means to $L$ classes and evaluate the performance. Simple hard clustering strategy can not adapt to arbitrary images with different number of classes and does not guarantee local smoothness, these two factors have large penalty on the evaluation metrics as shown in the first and second rows of Table \ref{table:bsds-ablation}. Therefore, we argue that our strategy to use eigenvectors for multi-class segmentation is very effective (as explained in Section 4.2). 

\subsubsection{Edge information} 
The improvement using MCG as the baseline is a small margin (i.e., $.1$ODS) compared with cases of using the other two methods as the baselines. In fact, MCG uses SE to detect edges while it also sharpens edges. Nevertheless, we observe MCG sometimes sharpens irrelevant edges as well, such that the sharpened noisy edges will have a large penalization through pairwise potentials against unary potentials in our multi-class segmentation procedure, leading to undesirable results. Figure \ref{fig:mcg_se_compare} illustrates this situation. The above results indicate that the proposed method relies less on strong edge information compared with MCG.

Additionally, since the proposed method relies less on edges, one potential weakness of the graph partitioning procedure could result in the fragmentation of homogeneous regions, which decreases the precision of the boundary detection. We compare the boundary precision-recall curve in Figure \ref{fig:edges}, from which we can see that the proposed method maintains nearly the same precision as the baseline methods, i.e., MCG and SE+ucm (though negligible 0.03 decrease for SE+ucm).

\subsubsection{Strengths and limitations} 
The proposed method is effective in discovering complex image knowledge among regions from challenging natural images and segmenting objects even when objects have weak boundaries. The proposed method is significantly better than MCG in those samples shown in Figure \ref{fig:comare_to_mcg}(left). However, we found that the proposed method is not that effective at images without definite objects, because our graph design emphasizes the high-level discriminative image knowledge of objects against the background. MCG outperforms ours in those samples (see Figure \ref{fig:comare_to_mcg}(right)).

\section{Conclusions}

In this paper, we present a fast yet accurate image segmentation method, which is a novel re-examination of spectral clustering based image segmentation for unsupervised image segmentation.
We construct an image region graph with both local and global connections based on simple but effective high-level cues, and formulate the graph partitioning as a simple generalized eigenvector system. The high quality graph partitions are used to compute effective unary potentials of Pylon model for multi-class image segmentation. Extensive experiments, on the BSDS500 benchmark, large-scale PASCAL VOC and COCO datasets, show that the proposed method achieves significantly faster speed and competitive performance when it is compared to state-of-the-art segmentation methods.

\section{Acknowledgement} 
This work was partially supported by the National Natural Science Foundation of China under Grants U1605252, 61472334, and 61571379.


\bibliography{reference}

\begin{thebibliography}{10}
\expandafter\ifx\csname url\endcsname\relax
  \def\url#1{\texttt{#1}}\fi
\expandafter\ifx\csname urlprefix\endcsname\relax\def\urlprefix{URL }\fi
\expandafter\ifx\csname href\endcsname\relax
  \def\href#1#2{#2} \def\path#1{#1}\fi

\bibitem{felzenszwalb2004efficient}
P.~F. Felzenszwalb, D.~P. Huttenlocher, Efficient graph-based image
  segmentation, International Journal of Computer Vision 59~(2) (2004)
  167--181.

\bibitem{boykov2001fast}
Y.~Boykov, O.~Veksler, R.~Zabih, Fast approximate energy minimization via graph
  cuts, Transactions on Pattern Analysis and Machine Intelligence 23~(11)
  (2001) 1222--1239.

\bibitem{peng2013survey}
B.~Peng, L.~Zhang, D.~Zhang, A survey of graph theoretical approaches to image
  segmentation, Pattern Recognition 46~(3) (2013) 1020--1038.

\bibitem{kim2013learning}
T.~H. Kim, K.~M. Lee, S.~U. Lee, Learning full pairwise affinities for spectral
  segmentation, Transactions on Pattern Analysis and Machine Intelligence
  35~(7) (2013) 1690--1703.

\bibitem{shi2015framework}
X.~Shi, Z.~Guo, Z.~Lai, Y.~Yang, Z.~Bao, D.~Zhang, A framework of joint graph
  embedding and sparse regression for dimensionality reduction, IEEE
  Transactions on Image Processing 24~(4) (2015) 1341--1355.

\bibitem{shi2000normalized}
J.~Shi, J.~Malik, Normalized cuts and image segmentation, Transactions on
  Pattern Analysis and Machine Intelligence 22~(8) (2000) 888--905.

\bibitem{yu2015piecewise}
Y.~Yu, C.~Fang, Z.~Liao, Piecewise flat embedding for image segmentation, in:
  Proceedings of the International Conference on Computer Vision, 2015, pp.
  1368--1376.

\bibitem{yu2005segmentation}
S.~X. Yu, Segmentation induced by scale invariance, in: Proceedings of the
  Conference on Computer Vision and Pattern Recognition, 2005, pp. 444--451.

\bibitem{maire2013progressive}
M.~Maire, S.~X. Yu, Progressive multigrid eigensolvers for multiscale spectral
  segmentation, in: Proceedings of the International Conference on Computer
  Vision, 2013, pp. 2184--2191.

\bibitem{cour2005spectral}
T.~Cour, F.~Benezit, J.~Shi, Spectral segmentation with multiscale graph
  decomposition, in: Proceedings of the Conference on Computer Vision and
  Pattern Recognition, 2005, pp. 1124--1131.

\bibitem{arbelaez2011contour}
P.~Arbelaez, M.~Maire, C.~Fowlkes, J.~Malik, Contour detection and hierarchical
  image segmentation, Transactions on Pattern Analysis and Machine Intelligence
  33~(5) (2011) 898--916.

\bibitem{arbelaez2014multiscale}
P.~Arbelaez, J.~Pont-Tuset, J.~Barron, F.~Marques, J.~Malik, Multiscale
  combinatorial grouping, in: Proceedings of the Conference on Computer Vision
  and Pattern Recognition, 2014, pp. 328--335.

\bibitem{chenscale}
Y.~Chen, D.~Dai, J.~Pont-Tuset, L.~Van~Gool, Scale-aware alignment of
  hierarchical image segmentation, in: Proceedings of the IEEE Conference on
  Computer Vision and Pattern Recognition, 2016, pp. 364--372.

\bibitem{belkin2003laplacian}
M.~Belkin, P.~Niyogi, Laplacian eigenmaps for dimensionality reduction and data
  representation, Neural computation 15~(6) (2003) 1373--1396.

\bibitem{ng2002spectral}
A.~Y. Ng, M.~I. Jordan, Y.~Weiss, et~al., On spectral clustering: Analysis and
  an algorithm, in: Advances in Neural Information Processing Systems, Vol.~14,
  2001, pp. 849--856.

\bibitem{shi2014face}
X.~Shi, Y.~Yang, Z.~Guo, Z.~Lai, Face recognition by sparse discriminant
  analysis via joint l 2, 1-norm minimization, Pattern Recognition 47~(7)
  (2014) 2447--2453.

\bibitem{lempitsky2011pylon}
V.~Lempitsky, A.~Vedaldi, A.~Zisserman, Pylon model for semantic segmentation,
  in: Advances in Neural Information Processing Systems, 2011, pp. 1485--1493.

\bibitem{everingham2010pascal}
M.~Everingham, L.~Van~Gool, C.~K. Williams, J.~Winn, A.~Zisserman, The pascal
  visual object classes (voc) challenge, International journal of computer
  vision 88~(2) (2010) 303--338.

\bibitem{lin2014microsoft}
T.-Y. Lin, M.~Maire, S.~Belongie, J.~Hays, P.~Perona, D.~Ramanan,
  P.~Doll{\'a}r, C.~L. Zitnick, Microsoft coco: Common objects in context, in:
  European conference on computer vision, Springer, 2014, pp. 740--755.

\bibitem{yu2003multiclass}
S.~X. Yu, J.~Shi, Multiclass spectral clustering, in: Transactions on Pattern
  Analysis and Machine Intelligence, 2003, pp. 313--319.

\bibitem{wang2012affinity}
B.~Wang, Z.~Tu, Affinity learning via self-diffusion for image segmentation and
  clustering, in: Computer Vision and Pattern Recognition (CVPR), 2012 IEEE
  Conference on, 2012, pp. 2312--2319.

\bibitem{zhang2010diffusion}
J.~Zhang, J.~Zheng, J.~Cai, A diffusion approach to seeded image segmentation,
  in: Computer Vision and Pattern Recognition (CVPR), 2010 IEEE Conference on,
  2010, pp. 2125--2132.

\bibitem{rother2004grabcut}
C.~Rother, V.~Kolmogorov, A.~Blake, Grabcut: Interactive foreground extraction
  using iterated graph cuts, in: ACM transactions on graphics (TOG), Vol.~23,
  2004, pp. 309--314.

\bibitem{grady2006random}
L.~Grady, Random walks for image segmentation, IEEE transactions on pattern
  analysis and machine intelligence 28~(11) (2006) 1768--1783.

\bibitem{li2016multiscale}
Y.~Li, X.~Feng, A multiscale image segmentation method, Pattern Recognition 52
  (2016) 332--345.

\bibitem{yin2017unsupervised}
S.~Yin, Y.~Qian, M.~Gong, Unsupervised hierarchical image segmentation through
  fuzzy entropy maximization, Pattern Recognition 68 (2017) 245--259.

\bibitem{de2014graph}
K.~J.~F. De~Souza, A.~de~Albuquerque~Ara{\'u}jo, Z.~K. do~Patroc{\'\i}nio,
  S.~J.~F. Guimar{\~a}es, Graph-based hierarchical video segmentation based on
  a simple dissimilarity measure, Pattern Recognition Letters 47 (2014) 85--92.

\bibitem{zhu2014constructing}
X.~Zhu, C.~Change~Loy, S.~Gong, Constructing robust affinity graphs for
  spectral clustering, in: Proceedings of the IEEE Conference on Computer
  Vision and Pattern Recognition, 2014, pp. 1450--1457.

\bibitem{li2012segmentation}
Z.~Li, X.-M. Wu, S.-F. Chang, Segmentation using superpixels: A bipartite graph
  partitioning approach, in: Proceedings of the Conference on Computer Vision
  and Pattern Recognition, 2012, pp. 789--796.

\bibitem{wang2015global}
X.~Wang, Y.~Tang, S.~Masnou, L.~Chen, A global/local affinity graph for image
  segmentation, Image Processing, IEEE Transactions on 24~(4) (2015)
  1399--1411.

\bibitem{isola2014crisp}
P.~Isola, D.~Zoran, D.~Krishnan, E.~H. Adelson, Crisp boundary detection using
  pointwise mutual information, in: European Conference on Computer Vision,
  2014, pp. 799--814.

\bibitem{xiaofeng2012discriminatively}
R.~Xiaofeng, L.~Bo, Discriminatively trained sparse code gradients for contour
  detection, in: Advances in Neural Information Processing Systems, 2012, pp.
  584--592.

\bibitem{donoser2014discrete}
M.~Donoser, D.~Schmalstieg, Discrete-continuous gradient orientation estimation
  for faster image segmentation, in: Proceedings of the Conference on Computer
  Vision and Pattern Recognition, 2014, pp. 3158--3165.

\bibitem{kim2014image}
S.~Kim, C.~D. Yoo, S.~Nowozin, P.~Kohli, Image segmentation using higher-order
  correlation clustering, Transactions on Pattern Analysis and Machine
  Intelligence 36~(9) (2014) 1761--1774.

\bibitem{dollar2015pami}
P.~Doll{\'a}r, C.~L. Zitnick, Fast edge detection using structured forests,
  Transactions on pattern analysis and machine intelligence 37~(8) (2015)
  1558--1570.

\bibitem{pont2017multiscale}
J.~Pont-Tuset, P.~Arbelaez, J.~T. Barron, F.~Marques, J.~Malik, Multiscale
  combinatorial grouping for image segmentation and object proposal generation,
  Transactions on pattern analysis and machine intelligence 39~(1) (2017)
  128--140.

\bibitem{taylor2013towards}
C.~J. Taylor, Towards fast and accurate segmentation, in: Proceedings of the
  Conference on Computer Vision and Pattern Recognition, 2013, pp. 1916--1922.

\bibitem{shen2017multi}
W.~Shen, B.~Wang, Y.~Jiang, Y.~Wang, A.~Yuille, Multi-stage
  multi-recursive-input fully convolutional networks for neuronal boundary
  detection, Proceedings of the International Conference on Computer Vision
  (ICCV).

\bibitem{Man+17}
K.~Maninis, J.~Pont-Tuset, P.~Arbel\'{a}ez, L.~V. Gool, Convolutional oriented
  boundaries: From image segmentation to high-level tasks, IEEE Transactions on
  Pattern Analysis and Machine Intelligence (TPAMI).

\bibitem{shen2015deepcontour}
W.~Shen, X.~Wang, Y.~Wang, X.~Bai, Z.~Zhang, Deepcontour: A deep convolutional
  feature learned by positive-sharing loss for contour detection, in:
  Proceedings of the IEEE Conference on Computer Vision and Pattern
  Recognition, 2015, pp. 3982--3991.

\bibitem{zhang2016semicontour}
Z.~Zhang, F.~Xing, X.~Shi, L.~Yang, Semicontour: A semi-supervised learning
  approach for contour detection, in: Proceedings of the Conference on Computer
  Vision and Pattern Recognition, 2016, pp. 251--259.

\bibitem{shen2016object}
W.~Shen, K.~Zhao, Y.~Jiang, Y.~Wang, Z.~Zhang, X.~Bai, Object skeleton
  extraction in natural images by fusing scale-associated deep side outputs,
  in: Proceedings of the IEEE Conference on Computer Vision and Pattern
  Recognition, 2016, pp. 222--230.

\bibitem{long2015fully}
J.~Long, E.~Shelhamer, T.~Darrell, Fully convolutional networks for semantic
  segmentation, in: Proceedings of the IEEE Conference on Computer Vision and
  Pattern Recognition, 2015, pp. 3431--3440.

\bibitem{chen2016deeplab}
L.-C. Chen, G.~Papandreou, I.~Kokkinos, K.~Murphy, A.~L. Yuille, Deeplab:
  Semantic image segmentation with deep convolutional nets, atrous convolution,
  and fully connected crfs, arXiv preprint arXiv:1606.00915.

\bibitem{noh2015learning}
H.~Noh, S.~Hong, B.~Han, Learning deconvolution network for semantic
  segmentation, in: Proceedings of the IEEE International Conference on
  Computer Vision, 2015, pp. 1520--1528.

\bibitem{zhou2013texture}
H.~Zhou, J.~Zheng, L.~Wei, Texture aware image segmentation using graph cuts
  and active contours, Pattern Recognition 46~(6) (2013) 1719--1733.

\bibitem{LI2016317}
Z.~Li, G.~Liu, D.~Zhang, Y.~Xu, Robust single-object image segmentation based
  on salient transition region, Pattern Recognition 52 (2016) 317 -- 331.

\bibitem{ladicky2010graph}
L.~Ladicky, C.~Russell, P.~Kohli, P.~H. Torr, Graph cut based inference with
  co-occurrence statistics, in: European Conference on Computer Vision, 2010,
  pp. 239--253.

\bibitem{fano1961transmission}
R.~M. Fano, D.~Hawkins, Transmission of information: A statistical theory of
  communications, American Journal of Physics 29~(11) (1961) 793--794.

\bibitem{belkin2001laplacian}
M.~Belkin, P.~Niyogi, Laplacian eigenmaps and spectral techniques for embedding
  and clustering., in: Advances in Neural Information Processing Systems, 2001,
  pp. 585--591.

\bibitem{roweis2000nonlinear}
S.~T. Roweis, L.~K. Saul, Nonlinear dimensionality reduction by locally linear
  embedding, Science 290~(5500) (2000) 2323--2326.

\bibitem{tenenbaum2000global}
J.~B. Tenenbaum, V.~De~Silva, J.~C. Langford, A global geometric framework for
  nonlinear dimensionality reduction, Science 290~(5500) (2000) 2319--2323.

\bibitem{parzen1962estimation}
E.~Parzen, On estimation of a probability density function and mode, The annals
  of mathematical statistics (1962) 1065--1076.

\bibitem{cheng2015global}
M.~Cheng, N.~J. Mitra, X.~Huang, P.~H. Torr, S.~Hu, Global contrast based
  salient region detection, Transactions on Pattern Analysis and Machine
  Intelligence 37~(3) (2015) 569--582.

\bibitem{kim2014salient}
J.~Kim, D.~Han, Y.-W. Tai, J.~Kim, Salient region detection via
  high-dimensional color transform, in: Proceedings of the Conference on
  Computer Vision and Pattern Recognition, 2014, pp. 883--890.

\bibitem{yang2013exemplar}
J.~Yang, Y.-H. Tsai, M.-H. Yang, Exemplar cut, in: Proceedings of the
  International Conference on Computer Vision, 2013, pp. 857--864.

\bibitem{farabet2013learning}
C.~Farabet, C.~Couprie, L.~Najman, Y.~LeCun, Learning hierarchical features for
  scene labeling, Transactions on Pattern Analysis and Machine Intelligence
  35~(8) (2013) 1915--1929.

\bibitem{liu2015crf}
F.~Liu, G.~Lin, C.~Shen, Crf learning with cnn features for image segmentation,
  Pattern Recognition 48~(10) (2015) 2983--2992.

\bibitem{kolmogorov2004energy}
V.~Kolmogorov, R.~Zabin, What energy functions can be minimized via graph
  cuts?, Transactions on Pattern Analysis and Machine Intelligence 26~(2)
  (2004) 147--159.

\bibitem{comaniciu2002mean}
D.~Comaniciu, P.~Meer, Mean shift: A robust approach toward feature space
  analysis, Transactions on Pattern Analysis and Machine Intelligence 24~(5)
  (2002) 603--619.

\bibitem{ren2013image}
Z.~Ren, G.~Shakhnarovich, Image segmentation by cascaded region agglomeration,
  in: Proceedings of the Conference on Computer Vision and Pattern Recognition,
  2013, pp. 2011--2018.

\bibitem{li2015unsupervised}
Y.~Li, M.~Paluri, J.~M. Rehg, P.~Doll{\'a}r, Unsupervised learning of edges,
  in: Proceedings of the Conference on Computer Vision and Pattern Recognition,
  2016, pp. 1619--1627.

\end{thebibliography}

\end{document}